\documentclass{article}
\usepackage{ijcai26}

\usepackage{times}
\usepackage{soul}
\usepackage{url}
\usepackage[hidelinks]{hyperref}
\usepackage[utf8]{inputenc}
\usepackage[small]{caption}
\usepackage{graphicx}
\usepackage{amsmath}
\usepackage{amsthm}
\usepackage{booktabs}
\usepackage[switch]{lineno}
\usepackage{algorithm}
\usepackage{algorithmic}
\usepackage{amssymb}
\usepackage{amsthm}
\usepackage{amsmath}
\usepackage{mathrsfs}
\usepackage{yfonts}
\theoremstyle{definition}
\newtheorem{definition}{Definition} 
\newtheorem{theorem}{Theorem} 
\newtheorem{assumption}{Assumption} 
\DeclareMathOperator*{\argmax}{argmax}
\DeclareMathOperator*{\argmin}{argmin}

\title{Amortized Multi-Objective Optimization Across Tasks with\\Generative Solution Modeling}

\author{
Tingyang Wei$^1$\and
Jiao Liu$^1$\and
Abhishek Gupta$^{2}$\footnote{Corresponding author}\and
Chin Chun Ooi$^3$\and\\
Puay Siew Tan$^4$\And
Yew-Soon Ong$^{1,3}$\\
\affiliations
$^1$College of Computing and Data Science, Nanyang Technological University, Singapore\\
$^2$School of Mechanical Sciences, Indian Institute of Technology, Goa, India\\
$^3$Centre for Frontier AI Research, Agency for Science, Technology and Research, Singapore\\
$^4$Singapore Institute of Manufacturing Technology, Agency for Science, Technology and Research, Singapore\\
\emails
\{tingyang001\}@e.ntu.edu.sg,
\{jiao.liu, asysong\}@ntu.edu.sg,
abhishekgupta@iitgoa.ac.in,
\{ooicc, tan\_puay\_siew\}@a-star.edu.sg
}

\begin{document}

\maketitle

\begin{abstract}
Many real-world applications require solving families of expensive multi-objective optimization problems~(EMOPs) under varying operational conditions. This can be formulated as parametric expensive multi-objective optimization problems (P-EMOPs) where each task parameter defines a distinct optimization instance. Current multi-objective Bayesian optimization methods have been widely used for finding finite sets of Pareto optimal solutions for each task. 
However, P-EMOPs present a fundamental challenge: the continuous task parameter space can contain infinite distinct problems, each requiring separate expensive evaluations. 
To address this, we propose learning an inverse model to amortize the multi-objective optimization cost across the continuous task-preference space, enabling direct solution prediction for any query without the need for expensive re-evaluation.
This paper introduces a novel parametric multi-objective Bayesian optimizer that learns this inverse model by alternating between (1) generative solution sampling via conditional generative models and (2) acquisition-driven search leveraging inter-task synergies. 
This approach enables effective optimization across multiple tasks and finally achieves direct solution prediction for unseen parameterized EMOPs without re-evaluations. We theoretically justify the faster convergence by leveraging inter-task synergies through task-aware Gaussian processes. Based on that, empirical studies in synthetic and real-world benchmarks further verify the effectiveness of the proposed parametric optimizer.
\end{abstract}

\section{Introduction}

Expensive multi-objective optimization problems (EMOPs) are crucial in black-box optimization, aiming to identify optimal trade-off solutions among multiple conflicting criteria under limited evaluation budgets~\cite{parego-1,pareto-n}. 
Due to their focus on conflicting objective functions and costly evaluations, EMOP formulations have been applied to a plethora of real-world applications encompassing materials discovery~\cite{beichen2024}, drug discovery~\cite{moo-gflownet}, and robotics~\cite{ICRA-1,ICRA-2}.

To efficiently identify trade-off solutions, surrogate modeling has become essential for reducing the cost of evaluating expensive objectives. 
Among these, multi-objective Bayesian optimization (MOBO) stands out as the leading strategy. 
MOBO methods have been widely utilized for searching finite sets of Pareto Sets by scalarizing multiple objectives into single objectives~\cite{parego-1,parego-2} or by optimizing indicators such as expected Hypervolume improvement~\cite{ehvi-1,ehvi-3} and uncertainty reduction~\cite{uncertainty-1,uncertainty-2,uncertainty-3,uncertainty-4}.
To move beyond these finite solution-set approximations and directly model the entire Pareto front, recent studies have extended MOBO to continuous Pareto‑manifold learning methods~\cite{moo-psl,CDM-PSL,PPSL-AAAI}.

However, in many real-world scenarios, optimization problems emerge not as isolated tasks but as families of related tasks under varying operational conditions. 
For example, turbine components must be optimized across different wind speeds~\cite{moo-gamut}, and robotic control policies must be optimized for diverse damage conditions~\cite{nature}. 
These cases lead to fundamental challenges: the continuous range of operational conditions forms infinitely many related optimization problems, each requiring separate, costly evaluations for current MOBO methods. 
As operational conditions frequently shift~\cite{moo-context-ICRA}, repeatedly performing full optimizations to evaluate trade-offs across multiple scenarios becomes prohibitively expensive and computationally infeasible. 
Therefore, tackling families of related EMOPs urges new approaches capable of amortizing the efforts of repeated multi-objective optimizations across the operational condition space into a lightweight framework.
To our best knowledge, little existing work has systematically addressed this parametric multi-task setting. Some related works are discussed in Appendix E.

This paper proposes to amortize the work of potential repeated multi-objective optimization across infinite tasks via a single inverse generative model pass, introducing a novel parametric multi-objective Bayesian optimizer.
Our approach alternates between two complementary phases: (1) generative solution sampling via conditional generative models, which learns high-quality solution distributions conditioned on task parameters and preferences, and (2) acquisition-driven search guided by task-aware Gaussian processes, which explicitly leverages inter-task correlations to efficiently explore the solution space. 
This dual strategy not only can enhance optimization efficiency across multiple related tasks but also finally builds an inverse model for predicting optimized solutions to previously unseen EMOPs without expensive re-evaluations.
The major contributions of this paper can be summarized as follows:
\begin{itemize}   

  \item We propose a novel inverse generative model to predict optimized solutions to each task query within the continuous task parameter space, thereby addressing the challenge of solving families of EMOPs.
  \item We design an alternating optimization approach that curates the conditional generative solution sampling with acquisition-driven search, enabling both efficient multi-task optimization and inverse model building.
  \item We develop a Parametric Multi-Task Multi-Objective Bayesian Optimizer~(PMT-MOBO) with task-aware Gaussian Processes, demonstrating its faster convergence over single-task counterparts empirically and theoretically.

\end{itemize}

\section{Preliminaries}
\subsection{Expensive Multi-Objective Optimization}
We consider expensive multi-objective optimization problems (EMOPs) defined as follows:
\begin{equation}
    \min_{\mathbf{x} \in \mathcal{X}} \, F(\mathbf{x}) := \min_{\mathbf{x} \in \mathcal{X}} \, \bigl(f_1(\mathbf{x}),\, \dots,\, f_M(\mathbf{x})\bigr),
\end{equation}
where $\mathbf{x} \in \mathcal{X} \subseteq \mathbb{R}^D$ denotes the decision variable, and each black-box objective function $f_m(x): \mathcal{X} \to \mathbb{R}, m = 1, \dots, M$, can be costly to evaluate and exhibits trade-offs with other objective functions.
The goal of optimizing EMOP is to approximate the Pareto set of non-dominated solutions that balance these conflicting objectives under limited evaluation budgets.
\begin{definition}[\textbf{Pareto Dominance}]
    Solution $\mathbf{x}^{(a)} \in \mathcal{X}$ is considered Pareto dominate another solution $\mathbf{x}^{(b)} \in \mathcal{X}$ if $\forall i \in [M]$, $f_{i}(\mathbf{x}^{(a)}) \leq f_{i}(\mathbf{x}^{(b)})$ and $\exists\,i' \in [M]$ such that $f_{i'}(\mathbf{x}^{(a)}) < f_{i'}(\mathbf{x}^{(b)})$~\cite{branke2008multiobjective}.
\end{definition}
\begin{definition}[\textbf{Pareto Optimality}]
    Solution $\mathbf{x}^{*} \in \mathcal{X}$ is considered Pareto optimal or non-dominated if there exists no other candidate solutions that can dominate $\mathbf{x}^{*}$. Consequently, Pareto Set (PS) and Pareto Front (PF) contain all the set of Pareto optimal solutions and their corresponding objective vectors~\cite{branke2008multiobjective}.
\end{definition}

\subsection{Parametric Expensive Multi-Objective Optimization}
In many real-world scenarios, EMOPs must be solved repeatedly under varying environmental~\cite{moo-gamut} or operational conditions~\cite{moo-qd}. 
These variations can be represented using continuous \emph{task parameters} $\boldsymbol{\theta} \in \Theta \subseteq \mathbb{R}^V$, which defines a family of different EMOPs over the same decision space. 
This leads to the formulation of a parametric expensive multi-objective optimization problem (P-EMOP):
\begin{equation}
\left\{
\begin{aligned}
&\min_{\mathbf{x} \in \mathcal{X}} \, F(\mathbf{x}, \boldsymbol{\theta}) := \min_{\mathbf{x} \in \mathcal{X}} \, \bigl(f_1(\mathbf{x}, \boldsymbol{\theta}),\, \dots,\, f_M(\mathbf{x}, \boldsymbol{\theta})\bigr) \\
&\mathcal{M}(\boldsymbol{\theta}, \boldsymbol{\lambda}) \approx \arg\min_{\mathbf{x} \in \mathcal{X}} \, \mathfrak{s}_{\boldsymbol{\lambda}}\bigl(F(\mathbf{x}, \boldsymbol{\theta})\bigr),\; \forall \boldsymbol{\lambda}\in\Lambda\subset\mathbb{R}^{M}
\end{aligned}
\right.
\label{eq: P-EMOP}
\end{equation}
where each task parameter $\boldsymbol{\theta}$ defines a distinct expensive multi-objective problem task $F(\mathbf{x}, \boldsymbol{\theta})$. 
The ideal goal of P-EMOP is to search all the Pareto Sets across potentially infinite task parameters under limited evaluation budgets.
We address this by learning an inverse model $\mathcal{M}(\boldsymbol{\theta}, \boldsymbol{\lambda})$ that generates optimized solutions for any task parameter $\boldsymbol{\theta}$ and preference vector $\boldsymbol{\lambda}$ without expensive re-optimization.
This represents a shift from single-task optimization to learning generalized solution mappings across the continuous task parameter space.
Here, $\mathfrak{s}_{\boldsymbol{\lambda}}(\cdot):\mathbb{R}^M \rightarrow \mathbb{R}$ is a scalarization function characterized by preference vector $\boldsymbol{\lambda}$.
We incorporate $\boldsymbol{\lambda}$ into the inverse model since the Pareto Set generally lies on a continuous manifold for non-trivial multi-objective optimization~\cite{moo-manifold}, enabling $\mathcal{M}(\boldsymbol{\theta}, \boldsymbol{\lambda})$ to output a single solution vector rather than a solution set.

Although (\ref{eq: P-EMOP}) contains infinitely many tasks, these EMOPs share inter-task synergies due to their parametric relationship. 
Rather than treating tasks in isolation, we solve a finite sample $\{\boldsymbol{\theta}_1, \dots, \boldsymbol{\theta}_K\}$ as a multi-task optimization problem~\cite{wei}, assuming tasks belong to a coherent family sharing solution components or objective landscape regularities. 
This assumption, pioneered by multi-objective multifactorial optimization~\cite{moomfea} and commonly exploited in subsequent multi-task optimization works~\cite{mfea,moo-finv,moo-mfea2} and recent parametric multi-objective optimization studies~\cite{ppsl}, enables more efficient solution search across the task parameter space through inter-task synergies, thereby yielding higher-quality solutions for constructing the inverse model.

\subsection{Gaussian Processes for Multi-Objective Optimization}
In this paper, we adopt Gaussian Process (GP)~\cite{GPML} as surrogate models to efficiently solve expensive multi-objective optimization problems (EMOPs). 
A GP defines a distribution over functions, assuming that $f \sim \mathcal{GP}(\mu(\cdot), \kappa(\cdot, \cdot))$, where $\mu(\mathbf{x}) = \mathbb{E} [f(\mathbf{x})]$ is the mean function and $\kappa(\mathbf{x}, \mathbf{x}') = Cov[f(\mathbf{x}), f(\mathbf{x}')]$ is the kernel or covariance function.
Given a dataset $\{(\mathbf{x}^{(i)}, y^{(i)})\}_{i=1}^N$, where each observation is corrupted by Gaussian noise $y^{(i)} = f(\mathbf{x}^{(i)}) + \epsilon^{(i)}, \epsilon^{(i)} \sim \mathcal{N}(0, \sigma_{\epsilon}^2)$, the posterior at a test point $\mathbf{x}$ is a normal distribution $\mathcal{N}(\mu(\mathbf{x}), \sigma^2(\mathbf{x}))$ with:
\begin{equation}
\left\{
\begin{aligned}
    &\mu(\mathbf{x}) = \mathbf{k}^\top (\mathbf{K} + \sigma_\epsilon^2 \mathbf{I}_N)^{-1} \mathbf{y}, \\
    \sigma^2(\mathbf{x}) &= \kappa(\mathbf{x}, \mathbf{x}) - \mathbf{k}^\top (\mathbf{K} + \sigma_\epsilon^2 \mathbf{I}_N)^{-1} \mathbf{k},
\end{aligned}
\label{eq: GP}
\right.
\end{equation}
where $\mathbf{k} \in \mathbb{R}^N$ is the kernel vector between $\mathbf{x}$ and training inputs, $\mathbf{K} \in \mathbb{R}^{N \times N}$ is the kernel matrix with $\mathbf{K}{p,q} = \kappa(\mathbf{x}^{(p)}, \mathbf{x}^{(q)})$, and $\mathbf{y} \in \mathbb{R}^N$ collects the observed outputs.

In EMOPs, Gaussian Processes (GPs) serve as probabilistic surrogates for approximating costly objective functions, enabling uncertainty-aware exploration using acquisition functions. 
Following the theoretically sound multi-objective Bayesian optimization~(MOBO) framework in ~\cite{parego-2}, in this paper, we adopt independent GP models for each objective and apply a scalarization-based acquisition function. 
In each iteration, we randomly sample a preference vector $\boldsymbol{\lambda} \in \Lambda$ that defines a trade-off among objectives via the scalarization function $\mathfrak{s}_{\boldsymbol{\lambda}}(\cdot)$, and optimize a scalarized acquisition function score, defined as:
\begin{equation}
   \mathbf{x}^{(t)} = \argmax_{\mathbf{x} \in \mathcal{X}} \mathfrak{s}_{\boldsymbol{\lambda}}\bigl(\boldsymbol{\alpha}(\mathbf{x})\bigr),
   \label{eq:acquisition}
\end{equation}
where $\boldsymbol{\alpha}(\mathbf{x}) \in \mathbb{R}^M$ denotes the vector of acquisition function values for each objective.
We refer to this method as ST-MOBO, which serves both as the base module in our framework and as a single-task~(ST) baseline EMOP solver in our empirical study.

\begin{figure*}[t]
\centering
\includegraphics[width=0.88\textwidth]{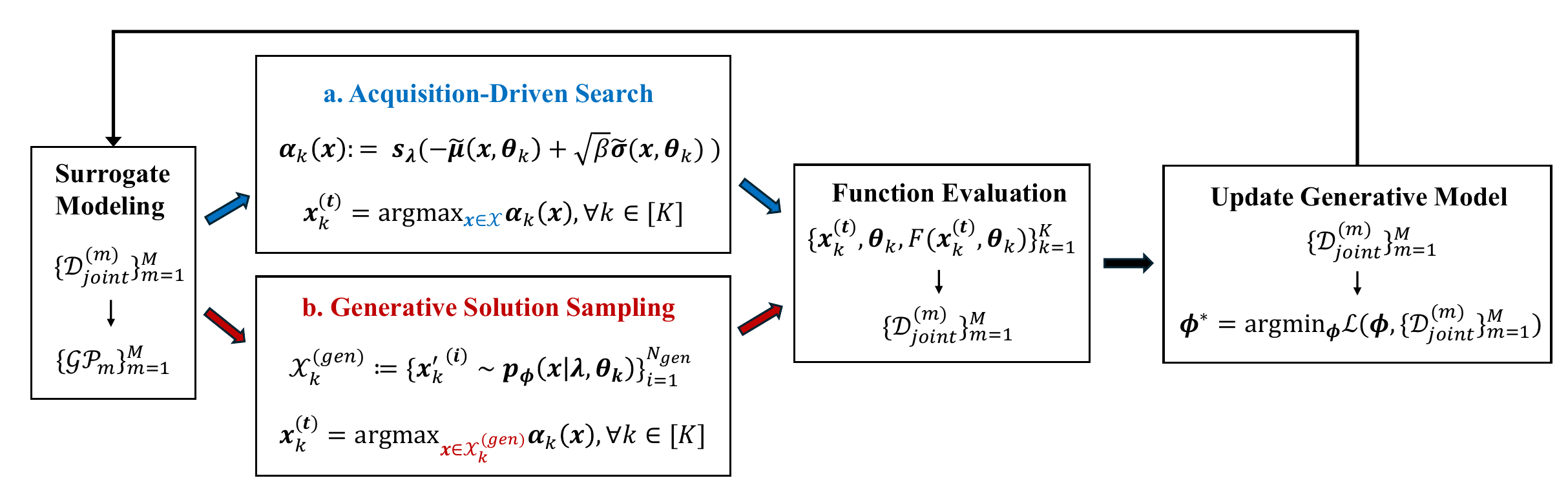}
\caption{The overall workflow of the proposed method. During the optimization process, solutions are sampled and selected in an alternating way between: a. acquisition-driven search and b. generative solution sampling.}
\label{fig: PMT-MOBO-X}
\end{figure*}

\section{Methodology}
\subsection{Overview}
To efficiently solve a family of EMOPs under limited evaluation budgets in (\ref{eq: P-EMOP}), we propose to build an inverse model by an alternating optimization framework that integrates task-aware surrogate modeling and conditional generative modeling, as shown in Figure~\ref{fig: PMT-MOBO-X}.
The former, instantiated via the Parametric Multi-Task Multi-Objective Bayesian Optimization~(PMT-MOBO) module, exploits inter-task relationships via task-aware GP to guide acquisition-driven search in a sample-efficient manner.
The latter, introduced as the conditional generative model, captures the distribution of high-performing solutions and enables task- and preference-aligned search in promising regions.
Both components are executed in an alternating loop, as illustrated in Figure~\ref{fig: PMT-MOBO-X}, enabling both sample-efficient multi-task search and finally achieving a high-quality inverse model.

\subsection{PMT-MOBO with Task-Aware GP}
We extend the baseline ST-MOBO to PMT-MOBO by incorporating task-aware GPs.
Each task-aware GP model receives both the decision variable $\mathbf{x}$ and the task parameter $\boldsymbol{\theta}$ as joint input.
Formally, for a given EMOP instance parametrized by $\boldsymbol{\theta}'$, the $m$-th objective is modeled as: \[f_m(\cdot, \boldsymbol{\theta}') \sim \mathcal{GP}_m\bigl(\widetilde\mu_{m}(\cdot, \boldsymbol{\theta}'), \widetilde\kappa_{m}((\cdot, \boldsymbol{\theta}'),(\cdot, \boldsymbol{\theta}'))\bigr), m\in[M]\] with mean functions and kernels defined over the joint input space $\mathcal{X} \times \Theta$.  
Based on these models, we extend the scalarized acquisition function defined in (\ref{eq:acquisition}) into follows:
\begin{equation}
   \mathbf{x}^{(t)}_k = \argmax_{\mathbf{x} \in \mathcal{X}} \mathfrak{s}_{\boldsymbol{\lambda}}\bigl(\boldsymbol{\widetilde\alpha}(\mathbf{x}, \boldsymbol{\theta}_k)\bigr),
   \label{eq: task-acquisition}
\end{equation}
where $\boldsymbol{\widetilde\alpha}(\mathbf{x}, \boldsymbol{\theta}_k) \in \mathbb{R}^M$ denotes the vector of task-aware acquisition function values for the EMOP task instance parameterized by $\boldsymbol{\theta}_k$. 
We refer to this method as PMT-MOBO.
For clarity, a comparative workflow of either ST-MOBO or PMT-MOBO is showcased in \textbf{Algorithm 2} of Appendix F.

By explicitly incorporating task parameters into the GP models, PMT-MOBO accounts for inter-task relationships during solution acquisition and model training. 
In the next subsection, we provide a theoretical analysis that quantifies its improved convergence compared to ST-MOBO.

\subsection{Theoretical Analysis of Faster Convergence}
For our theoretical analysis, we instantiate the acquisition functions $\boldsymbol{\alpha}(\mathbf{x})$ and $\boldsymbol{\widetilde\alpha}(\mathbf{x}, \boldsymbol{\theta}_k)$ using the Lower Confidence Bound (LCB)~\cite{andrea}. 
These LCB instantiations coincide with the acquisition rules employed in our practical ST-MOBO and PMT-MOBO algorithms, so the following regret analysis directly characterizes the convergence behavior of the implemented methods.
Specifically, for ST-MOBO, the acquisition function in (\ref{eq:acquisition}) becomes:
\begin{equation}
    \mathbf{x}^{(t)} = \argmax_{\mathbf{x} \in \mathcal{X}} \mathfrak{s}_{\boldsymbol{\lambda}}\bigl(-\boldsymbol{\mu}(\mathbf{x}) + \beta^{1/2}\boldsymbol{\sigma}(\mathbf{x})\bigr),
    \label{eq:ucb-mobo}
\end{equation}
where $\beta$ denotes a trade-off coefficient, and $\boldsymbol{\mu}(\cdot), \boldsymbol{\sigma}(\cdot) \in \mathbb{R}^M$ denote the vector of predictive means and standard deviations of $M$ GP models.
Correspondingly, for PMT-MOBO, the task-aware acquisition function in (\ref{eq: task-acquisition}) becomes:
\begin{equation}
    \mathbf{x}^{(t)}_k = \argmax_{\mathbf{x} \in \mathcal{X}} \mathfrak{s}_{\boldsymbol{\lambda}}\bigl(-\boldsymbol{\widetilde\mu}(\mathbf{x}, \boldsymbol{\theta}_k) + \beta^{1/2}\boldsymbol{\widetilde\sigma}(\mathbf{x}, \boldsymbol{\theta}_k)\bigr).
    \label{eq:pmt-mobo-ucb}
\end{equation}

For an EMOP with a total evaluation budget $T$, the goal is to obtain a set of solutions $\mathbf{X}_T \subset \mathcal{X}$ that approximates the true Pareto Set $\mathcal{X}^*$. 
Following the analysis technique~\cite{parego-2}, this goal can be quantified using the notion of \emph{Bayes Regret}, which evaluates the quality of the obtained solution set with respect to random preference vectors.

\begin{definition}[\textbf{Bayes Regret}]
Given a preference distribution \( p(\boldsymbol{\lambda}) \) over preference vectors, the Bayes regret of a solution set \( \mathbf{X}_T \) is defined as:
\[
\mathcal{R}_B(T) = \mathbb{E}_{\boldsymbol{\lambda} \sim p(\boldsymbol{\lambda})} \left[
\min_{\mathbf{x} \in \mathbf{X}_T} \mathfrak{s}_{\boldsymbol{\lambda}}\bigl(F(\mathbf{x})\bigr) -
\min_{\mathbf{x} \in \mathcal{X}^*} \mathfrak{s}_{\boldsymbol{\lambda}}\bigl(F(\mathbf{x})\bigr)
\right],
\label{eq: bayes_regret}
\]
where Pareto Set $\mathcal{X}^*$ includes all the Pareto-optimal solutions $\mathbf{x}_{\boldsymbol{\lambda}}^*=\argmin_{x\in \mathcal{X}} \mathfrak{s}_{\boldsymbol{\lambda}}(F(\mathbf{x})), \forall \boldsymbol{\lambda}\in\Lambda$.
\end{definition}

Deriving an upper bound on Bayes regret is non-trivial~\cite{hv-scalar}, since it measures the collective regret of the entire solution set rather than point-wise regret.
Therefore, a surrogate of Bayes regret is introduced, termed as \emph{Cumulative Regret}, which quantifies the total point-wise regret incurred across all evaluation rounds.

\begin{definition}[\textbf{Cumulative Regret}]
Given a set of sampled preference vectors $\{\boldsymbol{\lambda}^{(t)}\}_{t=1}^T$, the cumulative regret of a solution set $\mathbf{X}_T =\{\mathbf{x}^{(t)}\}_{t=1}^{T}$ is defined as:
\[
\mathcal{R}_C(T) = \sum_{t=1}^T \left[\mathfrak{s}_{\boldsymbol{\lambda}^{(t)}}(F(\mathbf{x}^{(t)})) -
\min_{\mathbf{x} \in \mathcal{X}^*} \mathfrak{s}_{\boldsymbol{\lambda}^{(t)}}(F(\mathbf{x}))\right].
\label{eq: cumulative_regret}
\]
\end{definition}

\begin{assumption}
For each objective $f_m$ and each task parameter $\boldsymbol{\theta}_k$, the
function $\mathbf{x}\mapsto f_m(\mathbf{x},\boldsymbol{\theta}_k)$ satisfies the
same regularity and GP assumptions as in Theorem~1
of~\cite{parego-2} (independent sub-Gaussian noise on a
compact domain, and scalarizations that are monotone and Lipschitz in all
objectives).
\end{assumption}


\begin{theorem}[Regret Bounds for ST-MOBO, Theorem 1 in~\cite{parego-2}]
Under Assumption 1, ST-MOBO satisfies:
\begin{equation}
  \mathbb{E}\bigl[\mathcal{R}_C(T)\bigr]
  = 
  \mathcal{O}\!\Bigl(
    J\,\sqrt{M^2\,T\,D\,\gamma_T\;\frac{\ln T}{\ln(1+\sigma^{-2})}}
  \Bigr),
  \label{eq: cumulative}
\end{equation}
and furthermore, its Bayes regret is controlled by:
\begin{equation}
  \mathbb{E}\bigl[\mathcal{R}_B(T)\bigr]
  \;\le\;
  \frac{1}{T}\,\mathbb{E}\bigl[\mathcal{R}_C(T)\bigr]
  \;+\;o(1).
  \label{eq: expected-bayes-regret}
\end{equation}
\end{theorem}

In Theorem 1, $\gamma_T$ denotes the largest \emph{maximum information gain}~(MIG) over all \(M\) objective functions:
\begin{equation}
\gamma_T \;=\;\max_{m \in [M]}\gamma_{T,m},
\label{eq: L-MIG}
\end{equation}
where $\gamma_{T,m}$ denotes the MIG for each objective function $f_m(\cdot)$, quantifying the
maximal uncertainty reduction about the objective function $f_m(\cdot)$ from observing $T$ solutions:
\begin{equation}
\gamma_{T,m}
\;=\;
\max_{{\{\mathbf{x}^{(t)}\}_{t=1}^T \subset \mathcal{X}}}
\;I\bigl(\{y^{(t)}_{m}\}_{t=1}^T ;\{f_m(\mathbf{x}^{(t)})\}_{t=1}^T\bigr),
\end{equation}
where $I$ stands for mutual information.
Although Theorem 1 is stated for a single EMOP instance, it applies equally well to each of the $K$ tasks in isolation.
For task $k$, let
\begin{equation}
  \gamma_{T,m,k}
  = 
  \max_{\{\mathbf{x}_k^{(t)}\}_{t=1}^T\subset\mathcal{X}}
    I\bigl(\{y_{m,k}^{(t)}\}_{t=1}^T;\{f_m(\mathbf{x}_k^{(t)},\boldsymbol{\theta}_k)\}_{t=1}^{T}\bigr)
    \label{eq: MIG}
\end{equation}
be the task-specific MIG of ST-MOBO and then \textbf{Theorem 1} gives the corresponding
\(\mathbb{E}[\mathcal{R}_C^{(k)}(T)]\) and \(\mathbb{E}[\mathcal{R}_B^{(k)}(T)]\) with \(\gamma_T\) replaced by \(\gamma_{T,k} = \max_m\gamma_{T,m,k}\).
With the same upper bound of Lipschitz constant $J$ and observation noise variance $\sigma^2$, the Bayes regret bound is determined primarily by per-task MIG \(\gamma_{T,k}\). 
A reduction in the MIG translates into a smaller upper bound on Bayes regret.
Thus, establishing that PMT-MOBO’s per-task MIG $\widetilde\gamma_{T,k}$ is smaller than that of ST-MOBO confirms its accelerated convergence.

\begin{theorem}[Tightened Regret Bounds for PMT-MOBO]
Under Assumption~1, consider a PMT-MOBO model where the task-aware kernel yields covariance values equivalent to the single-task kernel for any pair of inputs belonging to the same task. Let $\widetilde\gamma_{T,k} = \max_{m\in[M]}\widetilde\gamma_{T,m,k}$ be the largest MIG of PMT-MOBO across $M$ objectives over $T$ scalarized evaluations for task $k$.

Then, for all $m \in [M]$ and $k\in [K]$, it holds that
\begin{equation}
  \widetilde\gamma_{T,m,k} \;\le\; \gamma_{T,m,k},
\end{equation}
and accordingly $\widetilde\gamma_{T,k}\le\gamma_{T,k}$ for every $k\in[K]$.
Since the proof of Theorem~1 depends on each task only through its MIG term, the same regret bounds remain valid with $\gamma_{T,k}$ replaced by $\widetilde\gamma_{T,k}$, and consequently
\begin{equation}
  \mathbb{E}[\mathcal{R}_B^{\mathrm{PMT}}(T)]  \;\le\;  \mathbb{E}[\mathcal{R}_B(T)],
\end{equation}
i.e., PMT-MOBO achieves a tighter upper bound on Bayes regret than standard ST-MOBO.
\end{theorem}

\textbf{Theorem 2} establishes that PMT-MOBO achieves a tighter Bayes regret bound than ST-MOBO. The detailed proof (Appendix A) demonstrates that conditioning on observations from related tasks reduces the posterior covariance for the target task. Crucially, under the same regularity assumptions utilized in~\cite{parego-2}, one can ensure that the structural form of the original regret upper bound remains valid. Since this bound is a monotonically increasing function of the Maximum Information Gain (MIG), the inequality, $\widetilde\gamma_{T,k} \leq \gamma_{T,k}$ derived from the reduced posterior variance, translates to a strictly lower upper bound on the cumulative regret.

\begin{algorithm}[h]
\caption{Generative PMT-MOBO Framework}
\label{alg: pmt-mobo}
\textbf{Input}:  
  Tasks $\{\theta_k\}_{k=1}^K$, objectives $\{f_m\}_{m=1}^M$,  
  preference dist.\ $\Lambda$, generative model $\mathcal{G}_{\phi}$.\\
\textbf{Output}:  
  Final datasets $\{\mathcal{D}_k\}_{k=1}^K$

\begin{algorithmic}[1]
\FOR{$t=1$ \TO $T$}
  \STATE Sample $\lambda^{(t)} \sim \Lambda$
  \STATE \textbf{mode} $\leftarrow$ \texttt{ACQUISITION} or \texttt{GENERATIVE} (Round-Robin)
  \FOR{each task $\boldsymbol{\theta}_k$}
    \IF{\textbf{mode} = \texttt{ACQUISITION}}
      \STATE Select $\mathbf{x}_k^{(t)}$ using (\ref{eq: task-acquisition})
    \ELSE
      \STATE Sample $\{\mathbf{x}'^{(i)}_k\}_{i=1}^{N_{gen}} \sim p_\phi(\mathbf{x}|\boldsymbol{\lambda}^{(t)}, \boldsymbol{\theta}_k)$
      \STATE Select $\mathbf{x}_k^{(t)}$ using (\ref{eq: gen-acquisition})
    \ENDIF
    \STATE Evaluate $F(x_k,\theta_k)$ and append to $\mathcal{D}_k$
  \ENDFOR
  \STATE Jointly retrain \emph{all} task‑aware GPs on $\bigcup_k \mathcal{D}_k$
  \IF{\textbf{mode} = \texttt{GENERATIVE}}
    \STATE Extract elite subset from $\bigcup_k \mathcal{D}_k$
    \STATE Update $\mathcal{G}_{\phi}$ using $\mathcal{L}(\phi;\mathcal{D}^{(gen)})$ 
  \ENDIF
\ENDFOR
\STATE \textbf{return} $\{\mathcal{D}_k\}_{k=1}^K$
\end{algorithmic}
\end{algorithm}

\subsection{Conditional Generative Model}

Pure acquisition‐based search via PMT‐MOBO provides sample‐efficient guidance, but can steer sampling into low‐quality regions when $\mathcal{X}$ is large in (\ref{eq: task-acquisition}) and budgets are tight~\cite{cbas,LFI}.  
To complement this, we learn an elite‐solution distribution via a conditional generative model.  
As illustrated in Figure~\ref{fig: PMT-MOBO-X}, we let
\[
    \mathbf{x}' \sim p_{\boldsymbol{\phi}}\bigl(\mathbf{x}\bigm|\boldsymbol{\lambda},\boldsymbol{\theta}_k\bigr),
\]
where $\boldsymbol{\phi}$ denotes the model parameters, $\boldsymbol{\lambda}$ encodes the scalarization preference, and $\boldsymbol{\theta}_k$ specifies the current task $k$.  
We draw a pool of candidate solutions $\mathcal{X}_k^{(gen)}$ per task from this model, then score each with the acquisition function in (\ref{eq: task-acquisition}) and pick the best for evaluation as follows:
\begin{equation}
    \mathbf{x}_k^{(t)}=\argmax_{x\in\mathcal{X}_k^{(gen)}} \mathfrak{s}_{\boldsymbol{\lambda}}\bigl(-\boldsymbol{\widetilde\mu}(\mathbf{x}, \boldsymbol{\theta}_k) + \beta^{1/2}\boldsymbol{\widetilde\sigma}(\mathbf{x}, \boldsymbol{\theta}_k)\bigr),
    \label{eq: gen-acquisition}
\end{equation}
In this way, distribution‐aware sampling is seamlessly blended with uncertainty‐driven exploration.
We instantiate this generative model via either Variational Autoencoder~(VAE)~\cite{vae} or Denoising Diffusion Probabilistic Model~(DDPM)~\cite{ddpm} as follows.

\paragraph{Conditional Variational Autoencoder}
The dataset \(\mathcal{D}\) of elite solutions \(\mathbf{x}\) is collected with conditioning vectors \(\boldsymbol{c}=(\boldsymbol{\lambda},\boldsymbol{\theta}_k)\).  
We introduce a latent code \(z\) and parameterize
\begin{equation}
    p_{\phi}(\mathbf{x}|\boldsymbol{c})
  = \int p_{\phi_d}(\mathbf{x}| z,\boldsymbol{c})\,p(z|\boldsymbol{c})\,\mathrm{d}z.
\end{equation}
An encoder \(q_{\phi_e}(z|\mathbf{x},\boldsymbol{c})\) approximates the posterior, and a decoder \(p_{\phi_d}(\mathbf{x}| z,\boldsymbol{c})\) models the conditional likelihood.  
We learn \(\phi=(\phi_e,\phi_d)\) by maximizing the variational lower bound~\cite{vae,cvae} on \(\mathcal{D}\).
During generative solution sampling, candidate solutions \(\mathbf{x}'\) can be obtained from the learned distribution as follows:
\[
  z \sim p(z|\boldsymbol{c}), 
  \quad
  \mathbf{x}'\sim p_{\phi_d}(\mathbf{x}| z,\boldsymbol{c}).
\]

\begin{figure*}[t]
\centering
\includegraphics[width=0.95\textwidth]{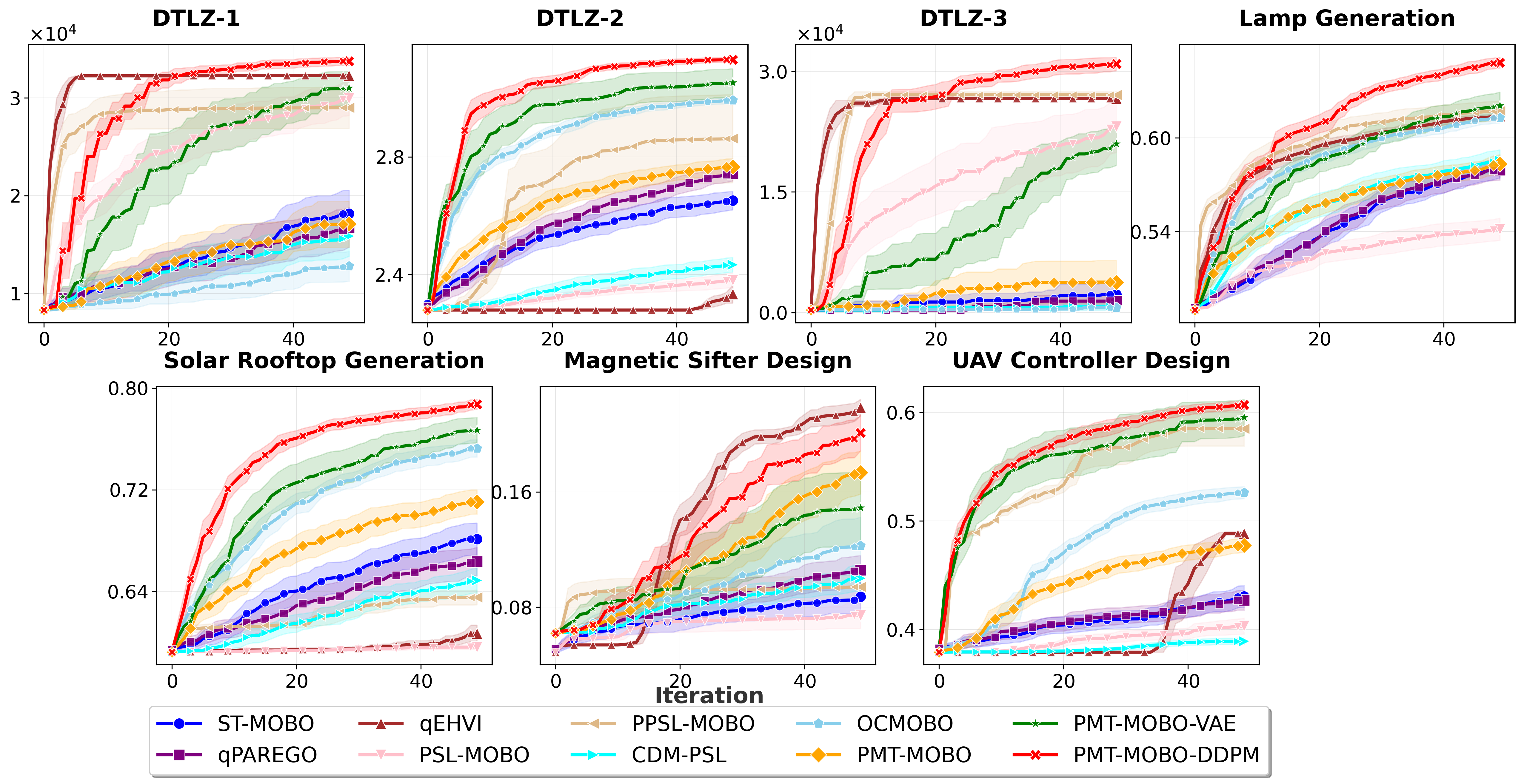}
\caption{Average Hypervolume results evaluated on sampled EMOP tasks in synthetic and real-world benchmarks.}
\label{fig: HV-Comparison}
\end{figure*}

\paragraph{Conditional Diffusion Model}
The dataset $\mathcal{D}$ of elite solutions $\mathbf{x}$ is collected with conditioning vectors $\boldsymbol{c}=(\boldsymbol{\lambda},\boldsymbol{\theta}_k)$.  
We model the data distribution through a forward diffusion process that gradually adds Gaussian noise over $\hat T$ timesteps:
\begin{equation}
    q(\mathbf{x}_t|\mathbf{x}_{t-1}) = \mathcal{N}(\mathbf{x}_t; \sqrt{1-\hat\beta_t}\mathbf{x}_{t-1}, \hat\beta_t\mathbf{I}),
\end{equation}
where $\{\hat \beta_t\}_{t=1}^{\hat T}$ follows a linear noise schedule. The forward process admits a closed-form expression:
\begin{equation}
    q(\mathbf{x}_t|\mathbf{x}_0) = \mathcal{N}(\mathbf{x}_t; \sqrt{\bar{\alpha}_t}\mathbf{x}_0, (1-\bar{\alpha}_t)\mathbf{I}),
\end{equation}
with $\bar{\alpha}_t = \prod_{s=1}^t (1-\hat \beta_s)$.
Thereafter, a neural net $\epsilon_\phi(\mathbf{x}_t, t, \boldsymbol{c})$ is trained to predict the noise added at timestep $t$ given the noisy solution $\mathbf{x}_t$ and conditioning $\boldsymbol{c}$. We learn $\phi$ by minimizing the denoising objective following~\cite{ddpm}.
The learned diffusion model parameterizes the conditional distribution of elite solutions as follows:
\begin{equation}
    p_{\phi}(\mathbf{x}|\boldsymbol{c}) = \int p_{\phi}(\mathbf{x}_0|\mathbf{x}_{\hat T}, \boldsymbol{c})\,p(\mathbf{x}_{\hat T})\,\mathrm{d}\mathbf{x}_{\hat T},\,
    p(\mathbf{x}_{\hat T}) = \mathcal{N}(0,\mathbf{I}),
\end{equation}
where $p_{\phi}(\mathbf{x}_0|\mathbf{x}_{\hat T}, \boldsymbol{c})$ represents the reverse denoising process implemented. 
During the generative solution sampling phase, candidate solutions $\mathbf{x}'$ are obtained as:
\[
  \mathbf{x}_{\hat T} \sim \mathcal{N}(0,\mathbf{I}), 
  \quad
  \mathbf{x}' \sim p_{\phi}(\mathbf{x}|\boldsymbol{c}).
\]

\subsection{Alternating Optimization Loop}
The complete framework alternates between two complementary search modes as illustrated in Figure~\ref{fig: PMT-MOBO-X}: acquisition-driven search via PMT-MOBO and generative solution sampling using the conditional generative model. Algorithm~\ref{alg: pmt-mobo} presents the detailed workflow alternating between \texttt{ACQUISITION} and \texttt{GENERATIVE} modes.
We alternate between generative sampling and acquisition-driven modes every iteration in a round-robin fashion.
In \texttt{GENERATIVE} mode, candidates are sampled from $p_\phi(\mathbf{x}|\boldsymbol{\lambda}^{(t)}, \boldsymbol{\theta}_k)$ and the best is selected using the acquisition function. 
In \texttt{ACQUISITION} mode, solutions are directly selected by optimizing the task-aware acquisition function.
Both update task-aware GPs after evaluation, while the generative model gets retrained during \texttt{GENERATIVE} iterations using $\mathcal{D}^{(gen)}$.

\paragraph{Elite Solution Collection}
We maintain datasets per task-preference pair $(\boldsymbol{\lambda}, \boldsymbol{\theta}_k)$, storing top $Q\%$ solutions by scalarization score $\mathfrak{s}_{\boldsymbol{\lambda}}(F(\mathbf{x}, \boldsymbol{\theta}_k))$~\cite{tpe}.
Preference vectors $\boldsymbol{\lambda}$ are generated via random sampling from $\Lambda$. 
Each elite solution $\mathbf{x}$ pairs with conditioning vector $\boldsymbol{c} = (\boldsymbol{\lambda}, \boldsymbol{\theta}_k)$ for generative model training.
More detailed settings can be found in Appendix C.3.

\begin{figure*}[th]
\centering
\includegraphics[width=0.95\textwidth]{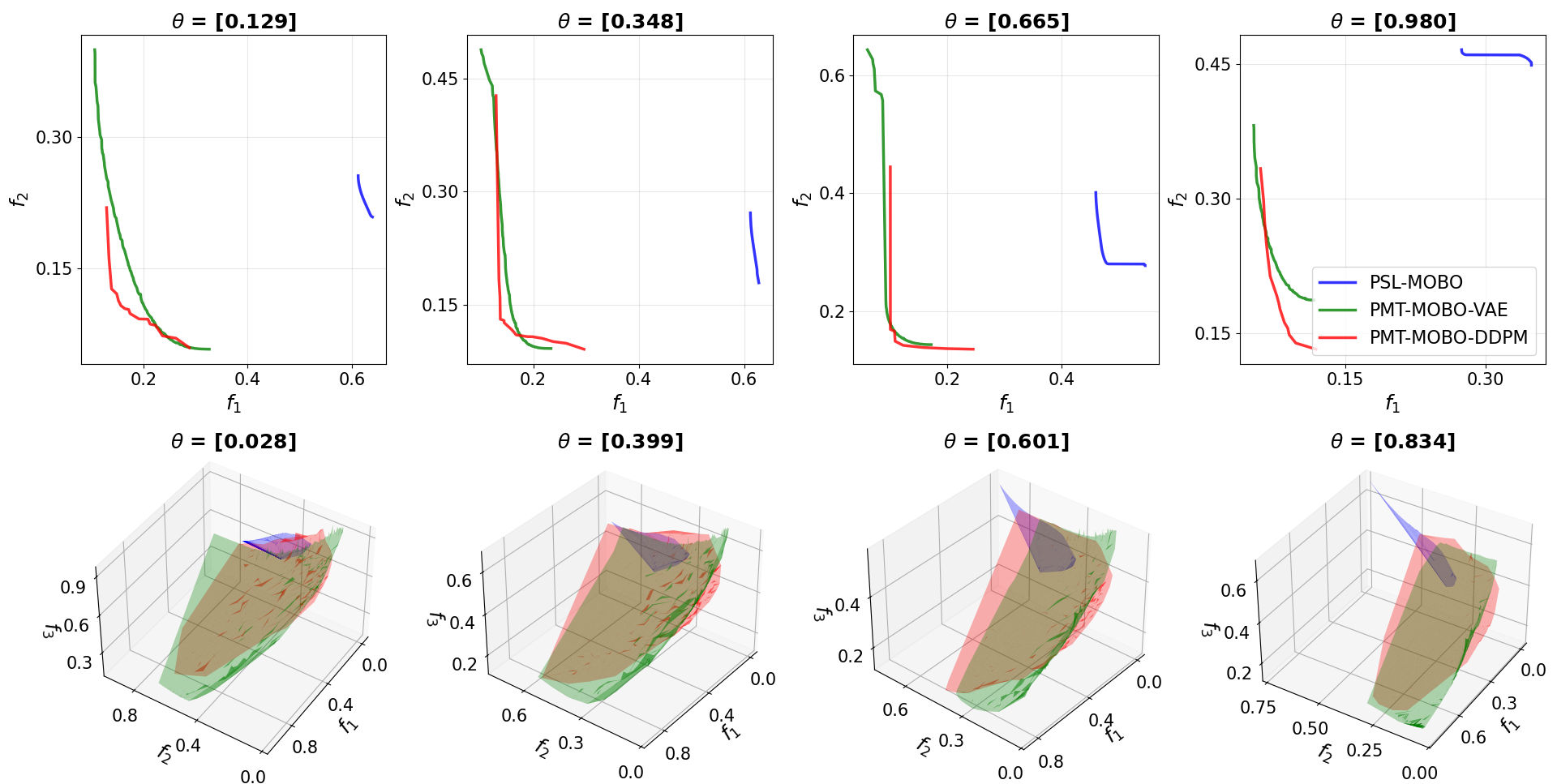}
\caption{Approximated Pareto Front for unseen EMOP tasks in solar rooftop generation (top) and lamp generation (bottom).}
\label{fig: PF-Comparison}
\end{figure*}

\begin{table*}[t]
\centering
\resizebox{1.00\textwidth}{!}{%
\begin{tabular}{l|l|l|l|l|l|l}
\hline
\textbf{Method} & \textbf{DTLZ-1} & \textbf{DTLZ-2} & \textbf{DTLZ-3} & \textbf{LAMP} & \textbf{SOLAR} & \textbf{UAV} \\
\hline
PSL-MOBO & 2.82e+04 (1.78e+03)$^{\approx -}$ & 1.90e+00 (1.30e-01)$^{- -}$ & 2.39e+04 (3.22e+03)$^{+ -}$ & 5.01e-01 (7.32e-02)$^{- -}$ & 3.54e-01 (4.06e-02)$^{- -}$ & 4.12e-01 (7.30e-03)$^{- -}$ \\
CDM-PSL & 1.92e+04 (3.85e+03)$^{- -}$ & 2.52e+00 (8.49e-02)$^{- -}$ & 1.83e+03 (1.31e+03)$^{- -}$ & 6.19e-01 (4.45e-02)$^{\approx -}$ & 6.54e-01 (3.11e-02)$^{- -}$ & 4.26e-01 (3.09e-02)$^{- -}$ \\
PPSL-MOBO & 2.24e+04 (1.10e+02)$^{- -}$ & 2.98e+00 (3.56e-02)$^{- -}$ & 2.67e+04 (6.08e+02)$^{+ -}$ & 6.12e-01 (1.05e-01)$^{\approx -}$ & 4.86e-01 (1.11e-01)$^{- -}$ & 5.24e-01 (1.71e-01)$^{- -}$ \\
PMT-MOBO-VAE & 2.88e+04 (2.57e+03) & 3.10e+00 (2.99e-02) & 1.78e+04 (7.28e+03) & 6.31e-01 (7.33e-02) & \textbf{7.93e-01 (1.56e-02)} & \textbf{6.08e-01 (1.88e-02)} \\
PMT-MOBO-DDPM & \textbf{3.47e+04 (4.85e+02)} & \textbf{3.13e+00 (1.41e-02)} & \textbf{3.20e+04 (1.98e+03)} & \textbf{6.70e-01 (5.82e-02)} & 7.65e-01 (1.30e-02) & 6.05e-01 (2.44e-02) \\
\hline
\end{tabular}
}
\caption{Inverse Model Generalization Performance on Unseen EMOP Tasks. Statistical significance is assessed using the Wilcoxon signed-rank test at the 95\% confidence level. Superscripts $^{s_1 s_2}$ indicate that the competitor is significantly worse ($-$), significantly better ($+$), or statistically similar ($\approx$) relative to \textbf{PMT-MOBO-VAE} ($s_1$) and \textbf{PMT-MOBO-DDPM} ($s_2$).}
\label{tab:inverse_model_performance}
\end{table*}

\section{Results}

We provide empirical experiments demonstrating our method's ability to build effective inverse models for P-EMOPs. 
Our experiments verify three key claims: (1) PMT-MOBO achieves faster convergence by exploiting inter-task synergies, (2) conditional generative models further improve solution quality, and (3) the built inverse model $\mathcal{M}(\boldsymbol{\theta}, \boldsymbol{\lambda})$ can directly predict optimized solutions for unseen EMOP tasks without additional evaluations. We evaluate all methods using the Hypervolume indicator~\cite{hv} across both seen and unseen EMOP tasks.
\paragraph{Benchmarks.}
We evaluate our methods on classical synthetic benchmarks: DTLZ-1, DTLZ-2, and DTLZ-3~\cite{DTLZ}, and real-world problems including lamp generation, solar rooftop generation~\cite{moo-gamut}, magnetic sifter design~\cite{extremo}, and UAV controller design~\cite{moo-gamut}. 
We modify these base EMOP problems so that each benchmark defines a family of related EMOPs, where $\boldsymbol{\theta}$ represents operational conditions (e.g., physical configurations of UAVs). 
More problem details are provided in the Appendix B.
\paragraph{Baselines.}
Our empirical studies include state-of-the-art single-task MOBO methods: ST-MOBO~\cite{parego-2}, qPAREGO~\cite{qpargeo}, qEHVI~\cite{ehvi-3}, PSL-MOBO~\cite{moo-psl}, and CDM-PSL~\cite{CDM-PSL}, a diffusion model based MOBO variant. Moreover, to evaluate performance in utilizing inter-task synergies, we compare against multi-task MOBO methods: OCMOBO~\cite{offline-contextual} and PPSL-MOBO~\cite{PPSL-AAAI}. Finally, we evaluate variants of our approaches that serve as ablation studies: PMT-MOBO (acquisition-driven only), PMT-MOBO-VAE, and PMT-MOBO-DDPM (full framework with different generative models).
\paragraph{Experimental Setup.}
We optimize $K = 8$ parameterized EMOPs per benchmark with uniformly sampled task parameters. 
Each task uses 20 random initial solutions and the budget $T = 50$, repeated over $U = 20$ independent runs. 
We set $Q = 10$ for elite solution extraction following~\cite{tpe} and $\beta$ follows~\cite{beta-schedule}. Additional implementation details are in the Appendix C.

\paragraph{Discussions.}
Figure~\ref{fig: HV-Comparison} and Table~\ref{tab:inverse_model_performance} demonstrate our framework's effectiveness on synthetic and real-world benchmarks. PMT-MOBO-VAE and PMT-MOBO-DDPM generally show superior or comparable results compared to state-of-the-art single-task and multi-task MOBO baselines in discovering high-quality Pareto solutions across distinct tasks. 
The exception in magnetic sifter motivates us to further explore the integration of our methods with EHVI based acquisition functions so that the proposed framework can be more flexibly adapt to diverse applications.
Table~\ref{tab:inverse_model_performance} further shows that our inverse models generalize well to unseen EMOP tasks, outperforming the single-task and multi-task inverse models constructed by PSL-MOBO and PPSL-MOBO. With task-aware kernel and the alternating optimization loop, our PMT-MOBO-DDPM's inverse model can also outperform the diffusion model of CDM-PSL across all the problems. To evaluate this generalization capability, we extract the trained inverse models after optimization and apply them to $W=100$ completely new tasks. For each unseen task, we query the inverse model with $S$ uniformly sampled preference vectors ($S = 100$ for bi-objective, $S = 1000$ for tri-objective) to generate predicted solutions, then compute the Hypervolume of the resulting solution set. 
\paragraph{Competitive Optimization Performance.} Figure~\ref{fig: HV-Comparison} and Table~5 in Appendix D.1 summarize the Hypervolume results across all benchmarks. Wilcoxon signed-rank tests (at 95\% confidence) demonstrate that our proposed methods significantly outperform the baselines in most scenarios. Specifically, against the eight competing methods across all tasks (56 pairwise comparisons), PMT-MOBO-VAE attains a win/tie/loss record of 45/8/3, effectively dominating difficult tasks such as LAMP and SOLAR. The PMT-MOBO-DDPM further enhances optimization capability, achieving a record of 55/0/1, statistically surpassing all baselines in every benchmark except for the low-dimensional magnetic sifter design.
\paragraph{Effective Inverse Model Generalization.} Table~\ref{tab:inverse_model_performance} confirms that our alternating optimization strategy yields robust inverse models for unseen tasks. PMT-MOBO-DDPM demonstrates superior generalization across the majority of benchmarks, achieving the highest Hypervolume on DTLZ-1, DTLZ-2, DTLZ-3, and LAMP generation. Notably, it statistically outperforms both PSL-MOBO and PPSL-MOBO baselines on every tested problem instance. PMT-MOBO-VAE excels on the remaining real-world engineering tasks, with the best performance on SOLAR and UAV controller design, though it struggles with the highly multimodal landscape of DTLZ-3 compared to the DDPM variant. We omit the magnetic sifter design set due to the prohibitive computational cost of evaluating hundreds of distinct tasks. To further validate these metrics, Figure~\ref{fig: PF-Comparison} visualizes the approximated Pareto Fronts. Both PMT-MOBO variants discover fronts with significantly better convergence and coverage than the PSL-MOBO baseline. Consistent with Table~\ref{tab:inverse_model_performance}, PMT-MOBO-DDPM generally provides the broadest coverage, while PMT-MOBO-VAE offers smoother approximations, illustrating a potential trade-off between the expansive generation of diffusion models and the structured latent embedding of VAEs.


\paragraph{Ablation Studies and Sensitivity Analysis.}
To validate the necessity of our alternating optimization strategy, we compared the full framework against decoupled baselines (pure acquisition search and pure generative sampling).
The results, detailed in Appendix D.3, confirm that neither component is sufficient in isolation. 
Specifically, generative sampling enhances the diversity of candidate generation, while the alternating acquisition search lowers the variance inherent in pure generative sampling, resulting in robust convergence.
Figure~\ref{fig: HV-Comparison} also demonstrates the superior performance of PMT-MOBO-VAE and PMT-MOBO-DDPM compared to PMT-MOBO, showcasing the value of integrating generative modeling. 

Furthermore, our sensitivity analysis on the elite ratio parameter $Q$ reveals a critical trade-off: too large a $Q$ reduces selective pressure, pushing the generative model to learn low-quality solutions, whereas too small a $Q$ incurs large variance and provides insufficient data to train a well-behaved model.
Empirically, $Q=10\%$ achieves the optimal balance across both VAE and DDPM variants (see Tables 6 and 7 in Appendix D.4).

\section{Conclusion}
We introduce a novel parametric multi-objective Bayesian optimizer that can amortize the computational effort of solving families of expensive EMOPs into a single inverse generative model. 
By alternating between acquisition-driven search and generative solution sampling, our method constructs a robust mapping that captures complex Pareto geometries, significantly improving generalization to unseen tasks and optimization efficiency, as verified by synthetic and real-world benchmarks. 
Theoretically, we showed that incorporating task-aware GPs yields tighter regret bounds than single-task counterparts. 
Future work will explore extensions to combinatorial problems and advanced acquisition strategies like EHVI to further broaden the impact of amortized optimization across tasks~\cite{pmto}.

\appendix

\section*{Acknowledgments}
This research is partly supported by the MTI under its AI Centre of Excellence for Manufacturing (AIMfg) (Award W25MCMF014), the Centre for Frontier AI Research (CFAR) under Agency for Science, Technology and Research (A*STAR), and the College of Computing and Data Science, Nanyang Technological University.
Abhishek Gupta is supported by the Ramanujan Fellowship from the Anusandhan National Research Foundation (ANRF), Government of India (Grant No. RJF/2022/000115).

\bibliographystyle{named}
\bibliography{ijcai26}

\newpage

\section{Proofs of Theorem 2}
\label{sec:proofs}
\begin{proof}
For task $k$, let
\begin{equation}
  \gamma_{T,m,k}
  = 
  \max_{\{\mathbf{x}_k^{(t)}\}_{t=1}^T\subset\mathcal{X}}
    I\bigl(\{y_{m,k}^{(t)}\}_{t=1}^T;\{f_m(\mathbf{x}_k^{(t)},\boldsymbol{\theta}_k)\}_{t=1}^{T}\bigr)
    \label{eq: newMIG}
\end{equation}
be the task-specific MIG of ST-MOBO and then Theorem 1 gives the corresponding
\(\mathbb{E}[\mathcal{R}_C^{(k)}(T)]\) and \(\mathbb{E}[\mathcal{R}_B^{(k)}(T)]\) with \(\gamma_T\) replaced by \(\gamma_{T,k} = \max_m\gamma_{T,m,k}\).

First, \text{w.l.o.g.}, we can rewrite (\ref{eq: newMIG}) in ST-MOBO as the following form of \emph{conditional information gain} about the $m$-th objective function with respect to any arbitrary parameterized EMOP instance of interest $k^* \in [K]$:
\begin{equation}
    \gamma_{T,m,k^*}
  = 
  \max_{\mathbf{X}_{T, k^*}\subset\mathcal{X}} I\bigl(\mathbf{y}_{m,k^* };\mathbf{f}_{m,k^* }|\mathcal{D}_{k^*}^{(m)}\bigr),
  \label{eq: per-UCB-MIG}
\end{equation}
where $\mathbf{X}_{T, k^*}$ is the solution set $\{\mathbf{x_{k^*}^{(t)}}\}_{t=1}^T$, $\mathbf{y}_{m,k^* }$ includes the noisy observations $\{y_{m,k^*}^{(t)}\}_{t=1}^T$, $\mathbf{f}_{m,k^* }$ contains the objective values $\{f_{m}(\mathbf{x}_{k^*}^{(t)}, \boldsymbol{\theta}_{k^*})\}_{t=1}^T$, and $\mathcal{D}_{k^*}^{(m)}$ is the collected dataset for the $m$-th objective function in the $k^*$-th EMOP task.
Similarly, MIG for PMT-MOBO can be represented as:
\begin{equation}
    \widetilde\gamma_{T,m,k^*}
  = 
  \max_{\mathbf{X}_{T,k^*}\subset\mathcal{X}} I\bigl(\mathbf{y}_{m,k^* };\mathbf{f}_{m,k^* }|\mathcal{D}_{\rm joint}^{(m)}\bigr),
  \label{eq: per-PMT-MIG}
\end{equation}
where the only distinction lies in that $\mathcal{D}_{\rm joint}^{(m)}$ contains data from all the $K$ tasks, $\bigcup_{k=1}^K\mathcal{D}_{k}^{(m)}$.

Since it is assumed that each objective function follows a Gaussian distribution, the \emph{conditional information gain} can be further computed as follows:
\begin{equation}
    I\bigl(\mathbf{y}_{m,k^* };\mathbf{f}_{m,k^* }|\mathcal{D}_{\rm joint}^{(m)}\bigr) = \frac{1}{2}\ln|\mathbf{I}+\sigma_{\epsilon}^{-2}\mathbf{K}^{(m)}_{{\rm pmt},k^*}|, 
    \label{eq: PMT-MIG}
\end{equation}
where $\mathbf{K}^{(m)}_{{\rm pmt},k^*}$ denotes the conditional covariance matrix for the parameterized task $k^*$, derived from the dataset $\mathcal{D}_{\rm joint}^{(m)}$, including the $m$-th objective values from all the tasks.
Next, we showcase how to compute $\mathbf{K}^{(m)}_{{\rm pmt},k^*}$.
Let $\pi_i, i\in[K]$ represents a rearrangement of values $\{1,2,\ldots,K\}$, then $\mathbf{K}^{(m)}_{\rm pmt}$ can be computed as follows:
\begin{equation}
\mathbf{K}_{\rm pmt}^{(m)} = 
\begin{bmatrix}
\mathbf{K}^{(m)}_{\pi_1,\pi_1} & \mathbf{K}^{(m)}_{\pi_1,\pi_2} & \cdots & \mathbf{K}^{(m)}_{\pi_1,\pi_K} \\
\mathbf{K}^{(m)}_{\pi_2,\pi_1} & \mathbf{K}^{(m)}_{\pi_2,\pi_2} & \cdots & \mathbf{K}^{(m)}_{\pi_2,\pi_K} \\
\vdots & \vdots & \ddots & \vdots \\
\mathbf{K}^{(m)}_{\pi_K,\pi_1} & \mathbf{K}^{(m)}_{\pi_K,\pi_2} & \cdots & \mathbf{K}^{(m)}_{\pi_K,\pi_K}
\end{bmatrix},
\label{eq: kernel_matrix}
\end{equation}
where each block matrix $\mathbf{K}^{(m)}_{i,j}$ can be denoted as follows:
\begin{equation}
    \mathbf{K}_{i,j}^{(m)}
=
\bigl[\;\widetilde\kappa_m\bigl((\mathbf{x}_i^{(t)},\boldsymbol{\theta}_i),\;(\mathbf{x}_j^{(t')},\boldsymbol{\theta}_j)\bigr)\bigr]_{t,t'=1}^{T}.
\end{equation}
Bearing the above in mind, we rewrite $\pi_K = k^*$, \begin{equation}
\mathbf{K}^{(m)}_{\setminus k^{*}} = 
\begin{bmatrix}
\mathbf{K}^{(m)}_{\pi_1,\pi_1} & \cdots & \mathbf{K}^{(m)}_{\pi_1,\pi_{K-1}} \\
\vdots & \ddots & \vdots \\
\mathbf{K}^{(m)}_{\pi_{K-1},\pi_1} & \cdots & \mathbf{K}^{(m)}_{\pi_{K-1},\pi_{K-1}}
\end{bmatrix},
\end{equation}
and $B = \begin{bmatrix} \mathbf{K}^{(m)}_{\pi_K,\pi_1} & \mathbf{K}^{(m)}_{\pi_K,\pi_2} & \cdots & \mathbf{K}^{(m)}_{\pi_K,\pi_{K-1}} \end{bmatrix}^T$, and thereby $\mathbf{K}^{(m)}_{\rm pmt}$ can be accordingly formulated as follows:
\begin{equation}
\mathbf{K}_{\rm pmt}^{(m)} = 
\begin{bmatrix}
\mathbf{K}^{(m)}_{\setminus k^{*}} & B \\
B^T & \mathbf{K}^{(m)}_{k^*,k^*}
\end{bmatrix}.
\label{eq: simplified_k_matrix}
\end{equation}
Consider the conditional Gaussian distribution~\cite{GPML,TLBO}, the conditional covariance matrix can be written as:
\begin{equation}
\begin{aligned}
\mathbf{K}^{(m)}_{{\rm pmt}, k^{*}} = \mathbf{K}^{(m)}_{k^*,k^*} - B^T(\mathbf{K}^{(m)}_{\setminus k^{*}} + \sigma_{\epsilon}^{-2}\mathbf{I})^{-1}B.
\end{aligned}
\label{eq: schur}
\end{equation}
Similarly, the \emph{conditional information gain} of ST-MOBO can be computed as follows:
\begin{equation}
    I\bigl(\mathbf{y}_{m,k^* };\mathbf{f}_{m,k^* }|\mathcal{D}_{k^*}^{(m)}\bigr) = \frac{1}{2}\ln|\mathbf{I}+\sigma_{\epsilon}^{-2}\mathbf{\mathbf{K}}^{(m)}_{{\rm st}, k^*}|,
    \label{eq: UCB-MIG}
\end{equation}
where it can be verified that $\mathbf{\mathbf{K}}^{(m)}_{{\rm st}, k^*} = \mathbf{K}^{(m)}_{k^*,k^*}$ due to the assumption $\widetilde\kappa_m\bigl((\mathbf{x},\boldsymbol{\theta}_k),(\mathbf{x}',\boldsymbol{\theta}_k)\bigr)
  = \kappa_m(\mathbf{x},\mathbf{x}'), \forall k \in [K]$.
Since the kernel matrix, $\mathbf{K}^{m}_{\rm pmt}$, in equation (\ref{eq: kernel_matrix}), is positive semi-definite~(PSD) and the matrix $\mathbf{K}^{(m)}_{\setminus k^{*}} + \sigma_{\epsilon}^{-2}\mathbf{I}$ in equation (\ref{eq: schur}) is invertible, then according to the \textit{Schur complement theorem}~\cite{schur}, it holds in equation (\ref{eq: schur}) that $\mathbf{K}^{(m)}_{k^*,k^*} - B^T(\mathbf{K}^{(m)}_{\setminus k^{*}} + \sigma_{\epsilon}^{-2}\mathbf{I})^{-1}B$ and $\mathbf{K}^{(m)}_{\setminus k^{*}} + \sigma_{\epsilon}^{-2}\mathbf{I}$ are also both PSD.
Considering the \textit{Minkowski determinant inequality} that for positive semi-definite matrices $C$ and $D$, we have $|C + D| \geq |C|+|D| \geq |C|$. After substituting $C$ and $D$ by $C = \mathbf{K}^{(m)}_{\setminus k^{*}} + \sigma_{\epsilon}^{-2}\mathbf{I}$ and $D = \mathbf{K}^{(m)}_{k^*,k^*} - B^T(\mathbf{K}^{(m)}_{\setminus k^{*}} + \sigma_{\epsilon}^{-2}\mathbf{I})^{-1}B$, respectively, the follows can be obtained:
\begin{equation}
\begin{aligned}
|\mathbf{I} + \sigma_{\epsilon}^{-2} \mathbf{K}^{(m)}_{k^{*}, k^{*}}| &= |\mathbf{I} + \sigma_{\epsilon}^{-2} \mathbf{K}^{(m)}_{{\rm st}, k^{*}}| \\ &\geq |\mathbf{I}+\sigma_{\epsilon}^{-2} {\mathbf{K}}^{(m)}_{{\rm pmt}, k^{*}}|.
\end{aligned}
\label{eq: geq_MIG}
\end{equation}
Thereafter, it can be deduced from (\ref{eq: per-UCB-MIG}, \ref{eq: per-PMT-MIG}, \ref{eq: PMT-MIG}, \ref{eq: UCB-MIG}, \ref{eq: geq_MIG}) that:
\begin{equation}
    \gamma_{T,m,k^*} \geq \widetilde\gamma_{T, m, k^*}, \forall m\in[M], \forall k^* \in [K].
\end{equation}
According to the definition of MIG, it also holds that:
\begin{equation}
    \gamma_{T,k^*} \geq \widetilde\gamma_{T, k^*}, \forall k^* \in [K].
\end{equation}
Considering that PMT-MOBO observing Assumption 1 as ST-MOBO does, the upper bounds of both $\mathbb{E}[\mathcal{R}_C^{\rm PMT}(T)]$ and $\mathbb{E}[\mathcal{R}_B^{\rm PMT}(T)]$ hold with only $\gamma_{T,k}$ replaced with $\widetilde\gamma_{T,k}$ in Theorem 1, the follows can be obtained:
\begin{equation}
    \mathbb{E}[\mathcal{R}_B^{\rm PMT}(T)] \leq \mathbb{E}[\mathcal{R}_B(T)],
\end{equation}
which completes the proof.
\end{proof}

\paragraph{Remark}
Theorem 2 relies on the covariance reduction property of the task-aware kernel structure.
Observing this property, the task-aware kernel reduces to the single-task kernel when the solution pair belongs to the same tasks. 
It should be noted that the adopted composite kernel, $\tilde{\kappa}((x, \theta), (x', \theta')) = \kappa_{dec}(x, x') \cdot \kappa_{task}(\theta, \theta')$, observes this theoretical assumption.
And this type of kernel can represent diverse kernel choices like composite kernel~\cite{offline-contextual} and multi-task kernel~\cite{swersky2013multi}, commonly seen in multi-task GP surrogates. 
The validity of this tighter regret bound is also empirically verified by the superior convergence shown across our benchmarks, which is detailed in Table \ref{tab:results} and Figure \ref{fig: HV-Comparison} comparing PMT-MOBO to ST-MOBO.
Crucially, the theoretical derivation of Theorem 2 introduces no additional requirements beyond those already established in Assumption 1. 
Since the noisy observations remain sub-Gaussian and the scalarization function remains monotone and Lipschitz, the structural validity of the original regret upper bound in Theorem 1 is preserved. 
Theorem 2 thereby demonstrates that under the same regularity conditions, the task-aware kernel can decrease the Maximum Information Gain (MIG), the core governing term of the regret bound, thereby tightening the bound without necessitating more restrictive assumptions.

Furthermore, it is acknowledged that we have not yet derived a formal regret bound for the alternating generative solution sampling, but only for the existing tightened regret bound for the task-aware GP.
The generative sampling acts as a heuristic proposal distribution to help mitigate the bias of pure acquisition search and escape local optima, bolstered by our empirical studies and ablation studies. Extending the theoretical analysis to unify both the multi-objective Bayesian optimization bounds and the generative sampling dynamics remains a non-trivial challenge. 
It is left as a potential direction for future work.

\section{Benchmark Definitions}
\label{sec:benchmarks}

\begin{table*}[h]
\centering
{\small
\setlength{\tabcolsep}{1mm}
\begin{tabular}{l|l|l|l|l|l|l|l}
\hline
\textbf{Benchmark}    & \textbf{DTLZ‑1}    & \textbf{DTLZ‑2}    & \textbf{DTLZ‑3}    & \textbf{LAMP}      & \textbf{SOLAR}  &   \textbf{MAGNETIC}   & \textbf{UAV}       \\
\hline
$D~(\mathcal{X}\subset\mathbb{R}^D)$             & 8                  & 8                  & 8                  & 9                  & 9                  &  3  & 12                 \\
$V~(\boldsymbol{\Theta}\subset\mathbb{R}^V)$       & 1                  & 1                  & 1                  & 1                  & 1                  & 2    & 2              \\
$M\bigl(F(\mathbf{x}): \mathcal{X}\xrightarrow{}\mathbb{R}^M\bigr)$        & 2                  & 2                  & 2                  & 3                  & 2        &3          & 2                  \\
\hline
\end{tabular}
}
\caption{Benchmark Problem Specifications: The dimensions of Solution Space $\mathcal{X}$, Task Space $\boldsymbol{\Theta}$, and Objective Space.}
\label{tab:benchmark_specs}
\end{table*}

\begin{table*}[h]
\centering
{\small
\setlength{\tabcolsep}{1mm}
\begin{tabular}{l|l|l|l}
\hline
\textbf{Variants} & \textbf{DTLZ-1} & \textbf{DTLZ-2} & \textbf{DTLZ-3} \\
\hline
PMT-MOBO (w/o \texttt{GENERATIVE}) & 1.709e+04 (4.696e+03) & 2.767e+00 (5.500e-02) & 3.785e+03 (5.397e+03) \\
PLAIN-VAE (w/o \texttt{ACQUISITION}) & 2.052e+04 (7.555e+03) & 2.933e+00 (2.315e-01) & 1.301e+04 (8.679e+03) \\
PMT-MOBO-VAE & \textbf{3.101e+04 (3.415e+03)} & \textbf{3.052e+00 (9.580e-02)} & \textbf{2.098e+04 (5.335e+03)} \\
\hline
\end{tabular}
}
\caption{Ablation Study of VAE Integration. The Alternating Loop (PMT-MOBO-VAE) consistently outperforms both pure Acquisition Search and pure Generative Sampling.}
\label{tab:ablation_vae}
\end{table*}

\begin{table*}[h]
\centering
{\small
\setlength{\tabcolsep}{1mm}
\begin{tabular}{l|l|l|l}
\hline
\textbf{Variants} & \textbf{DTLZ-1} & \textbf{DTLZ-2} & \textbf{DTLZ-3} \\
\hline
PMT-MOBO (w/o \texttt{GENERATIVE}) & 1.709e+04 (4.696e+03) & 2.767e+00 (5.500e-02) & 3.785e+03 (5.397e+03) \\
PLAIN-DDPM (w/o \texttt{ACQUISITION}) & 1.384e+04 (3.847e+03) & 3.052e+00 (3.970e-02) & 2.337e+03 (3.425e+03) \\
PMT-MOBO-DDPM & \textbf{3.374e+04 (8.354e+02)} & \textbf{3.131e+00 (1.620e-02)} & \textbf{3.093e+04 (1.673e+03)} \\
\hline
\end{tabular}
}
\caption{Ablation Study of DDPM Integration. The proposed PMT-MOBO-DDPM consistently achieves superior performance, particularly on deceptive landscapes (DTLZ-3) where plain generation fails.}
\label{tab:ablation_ddpm}
\end{table*}

In Table \ref{tab:benchmark_specs}, we specify the dimension of solution and task spaces for each benchmark problem as well as the number of objectives.
Note that, since the problem formulation we focus on is expensive multi-objective optimization problems, too large a solution or task space might easily incur a dimensionality curse for those surrogate models and generative models. Therefore, in this paper, we mainly focus on solving low-to-moderate P-EMOPs. We leave solving high-dimensional P-EMOPs as a future challenging direction.

\subsection{Synthetic Benchmarks}
We evaluate our approach on three classical multi-objective test problems from the DTLZ suite: DTLZ-1, DTLZ-2, and DTLZ-3~\cite{DTLZ}. 
These scalable benchmark functions are widely used for evaluating multi-objective optimization algorithms due to their well-understood theoretical properties and configurable Pareto Front geometries~\cite{DTLZ}.

To create parametric families of these problems, we introduce the task parameter $\boldsymbol{\theta}\in [0.8,1]$. 
The task parameter $\boldsymbol{\theta}$ controls the power values applied to the decision variables $\mathbf{x}$ in the objective evaluations, generating variations in the problem landscape in the form $F(\mathbf{x}^{\boldsymbol{\theta}})$.
This parameterization enables the exploration of related optimization instances that share a similar structure but exhibit different convergence characteristics and Pareto Front properties. 
To compare the multi-objective optimization performance, the reference point for each benchmark is set to $[200.0, 200.0]$, $[2.0, 2.0]$, and $[240.0, 240.0]$.

\subsection{Real-World Benchmarks}
We evaluate our framework on four real problem sets that demonstrate real applications across diverse domains. 
Each problem involves optimizing design variables $\mathbf{x}$ under varying operational conditions $\boldsymbol{\theta}$, where both represent meaningful physical quantities.

\paragraph{Lamp Generation} 
The design variables $\mathbf{x}$ encode the geometric structure of a three-arm symmetric desk lamp.
They include base dimensions, arm templates, and hand orientations that are rotated to create the complete fixture. 
The operational condition $\boldsymbol{\theta}$ stands for the desired focal point height, encompassing desk-level to floor-lamp configurations across a continuous range. 
We try to search the Pareto Sets among three objective functions: structural stability (minimizing wobbliness under perturbations), material efficiency (reducing the weight and cost), and illumination quality (maximizing light coverage at the target location).
To compare the multi-objective optimization performance, the reference point for this problem is set to $[1.0, 1.0, 1.0]$.

\paragraph{Solar Rooftop Generation} 
The design variables $\mathbf{x}$ stand for the elevation profile of interior grid points on a rooftop solar installation, allowing for generating complex geometries while maintaining fixed boundary conditions. 
The operational condition $\boldsymbol{\theta}$ captures the orientation of the building relative to the cardinal directions in a continuous rotational space.
Inhere, we consider a tri-objective optimization problem that balances energy harvest during morning hours (capturing early sunlight for peak demand periods) against evening energy collection (storing power for night-time usage).
To compare the multi-objective optimization performance, the reference point for this problem is set to $[1.0, 1.0]$.

\paragraph{UAV Controller Design} 
The design variables $\mathbf{x}$ stand for thrust commands for a twin-rotor UAV over multiple time steps.
They can be used to define the complete flight trajectory from initial state to target destination. The operational conditions $\boldsymbol{\theta}$ involve the vehicle's physical characteristics, including fuselage length and structural density, representing different UAV configurations. 
The controller must simultaneously minimize navigation accuracy (how close the vehicle gets to its target destination) and efficiency (conserving battery power for further flight duration). 
And this is a classic trade-off between precision and endurance in aerial robotics~\cite{bicopter}.
To compare the multi-objective optimization performance, the reference point for this problem is set to $[1.0, 1.0]$.

\paragraph{Magnetic Sifter Design} 
The design variables $\mathbf{x}$ are the geometric parameters of a biomedical microdevice, including channel gap width, magnetic element length, and structural thickness. 
The operational conditions $\boldsymbol{\theta}$ represent the magnetic properties of both cell types that need to be separated, varying across a continuous range of real values. 
The device should satisfy three conflicting objectives: purity (minimizing capture of unwanted cell types), selectivity (maximizing the separation quality between desired and undesired cells), and manufacturing efficiency (reducing device thickness to minimize fabrication cost)~\cite{extremo}. 
For real biomedical engineering, it is challengin that the same device design must be optimized for separating different patient samples or cell types with distinct magnetic characteristics from complex biological mixtures such as blood.
To compare the multi-objective optimization performance, the reference point for this problem is set to $[1.0, 1.0, 1.0]$.

\begin{table*}[t]
\centering
\resizebox{0.99\textwidth}{!}{%
\begin{tabular}{l|l|l|l|l|l|l|l}
\hline
\textbf{Method} & \textbf{DTLZ-1} & \textbf{DTLZ-2} & \textbf{DTLZ-3} & \textbf{LAMP} & \textbf{SOLAR} & \textbf{MAGNETIC} & \textbf{UAV} \\
\hline
ST-MOBO & 1.82e+04 (4.80e+03)$^{- -}$ & 2.65e+00 (6.30e-02)$^{- -}$ & 2.24e+03 (3.03e+03)$^{- -}$ & 5.80e-01 (1.20e-02)$^{- -}$ & 6.81e-01 (2.51e-02)$^{- -}$ & 8.72e-02 (2.15e-02)$^{- -}$ & 4.31e-01 (1.99e-02)$^{- -}$ \\
qPAREGO & 1.68e+04 (4.03e+03)$^{- -}$ & 2.74e+00 (5.30e-02)$^{- -}$ & 1.42e+03 (1.48e+03)$^{- -}$ & 5.79e-01 (1.30e-02)$^{- -}$ & 6.64e-01 (2.21e-02)$^{- -}$ & 1.06e-01 (2.05e-02)$^{- -}$ & 4.27e-01 (1.82e-02)$^{- -}$ \\
qEHVI & \textbf{3.23e+04 (5.06e+01)}$^{\approx -}$ & 2.33e+00 (6.20e-02)$^{- -}$ & 2.66e+04 (1.85e-02)$^{+ -}$ & 6.14e-01 (3.00e-03)$^{- -}$ & 6.07e-01 (1.29e-02)$^{- -}$ & \textbf{2.18e-01 (1.15e-02)}$^{+ +}$ & 4.89e-01 (2.60e-03)$^{- -}$ \\
PSL-MOBO & 3.00e+04 (3.52e+03)$^{\approx -}$ & 2.38e+00 (4.90e-02)$^{- -}$ & 2.31e+04 (6.44e+03)$^{\approx -}$ & 5.42e-01 (1.40e-02)$^{- -}$ & 5.96e-01 (3.60e-03)$^{- -}$ & 7.42e-02 (1.79e-02)$^{- -}$ & 4.03e-01 (1.46e-02)$^{- -}$ \\
CDM-PSL & 1.59e+04 (4.15e+03)$^{- -}$ & 2.43e+00 (5.04e-02)$^{- -}$ & 7.76e+02 (9.41e+02)$^{- -}$ & 5.85e-01 (1.39e-02)$^{- -}$ & 6.49e-01 (1.97e-02)$^{- -}$ & 1.00e-01 (1.82e-02)$^{- -}$ & 3.89e-01 (5.70e-03)$^{- -}$ \\
PPSL-MOBO & 3.20e+04 (1.72e+03)$^{\approx -}$ & 2.86e+00 (3.00e-01)$^{\approx -}$ & \textbf{2.71e+04 (7.89e+02)}$^{+ -}$ & 6.09e-01 (1.61e-02)$^{- -}$ & 6.35e-01 (1.15e-02)$^{- -}$ & 9.41e-02 (1.69e-02)$^{- -}$ & 5.85e-01 (3.19e-02)$^{\approx -}$ \\
OCMOBO & 1.28e+04 (3.04e+03)$^{- -}$ & 2.99e+00 (2.12e-02)$^{\approx -}$ & 6.20e+02 (6.29e+02)$^{- -}$ & 6.13e-01 (6.60e-03)$^{- -}$ & 7.53e-01 (1.36e-02)$^{- -}$ & 1.23e-01 (3.95e-02)$^{- -}$ & 5.26e-01 (1.09e-02)$^{- -}$ \\
\hline
PMT-MOBO & 1.71e+04 (4.70e+03)$^{- -}$ & 2.77e+00 (5.50e-02)$^{- -}$ & 3.78e+03 (5.40e+03)$^{- -}$ & 5.83e-01 (1.22e-02)$^{- -}$ & 7.10e-01 (1.90e-02)$^{- -}$ & 1.73e-01 (2.99e-02)$^{\approx -}$ & 4.78e-01 (1.27e-02)$^{- -}$ \\
PMT-MOBO-VAE & 3.10e+04 (3.41e+03) & \textbf{3.05e+00 (9.58e-02)} & 2.10e+04 (5.33e+03) & \textbf{6.21e-01 (1.73e-02)} & \textbf{7.67e-01 (2.01e-02)} & 1.49e-01 (5.07e-02) & \textbf{5.95e-01 (3.33e-02)} \\
PMT-MOBO-DDPM & \textbf{3.37e+04 (8.35e+02)} & \textbf{3.13e+00 (1.62e-02)} & \textbf{3.09e+04 (1.67e+03)} & \textbf{6.48e-01 (5.10e-03)} & \textbf{7.87e-01 (7.80e-03)} & \textbf{2.01e-01 (2.50e-02)} & \textbf{6.07e-01 (7.50e-03)} \\
\hline
\end{tabular}%
}
\caption{Average Hypervolume results (mean $\pm$ std) across synthetic and real-world benchmarks. Statistical significance is assessed using the Wilcoxon signed-rank test at the 95\% confidence level.
Superscripts $^{s_1 s_2}$ indicate that the competitor is significantly worse ($-$), significantly better ($+$), or statistically similar ($\approx$) relative to \textbf{PMT-MOBO-VAE} ($s_1$) and \textbf{PMT-MOBO-DDPM} ($s_2$).}
\label{tab:results}
\end{table*}
\begin{table*}[h]
\centering
{\small
\setlength{\tabcolsep}{1mm}
\begin{tabular}{l|l|l|l}
\hline
\textbf{Variants ($Q\%$)} & \textbf{DTLZ-1} & \textbf{DTLZ-2} & \textbf{DTLZ-3} \\
\hline
PMT-MOBO-VAE-S (5\%) & 2.523e+04 (4.644e+03) & 3.046e+00 (6.060e-02) & \textbf{2.256e+04 (7.200e+03)} \\
PMT-MOBO-VAE (10\%) & \textbf{3.101e+04 (3.415e+03)} & \textbf{3.052e+00 (9.580e-02)} & 2.098e+04 (5.335e+03) \\
PMT-MOBO-VAE-L (20\%) & 2.802e+04 (3.211e+03) & 2.995e+00 (7.380e-02) & 1.576e+04 (9.457e+03) \\
\hline
\end{tabular}%
}
\caption{Sensitivity Analysis of Elite Ratio ($Q\%$) for VAE. The base setting (10\%) is generally most robust, although PMT0-MOBO-VAE-S (5\%) excels on DTLZ-3 by avoiding local optimum overfitting but resulting in a larger variance.}
\label{tab:sensitivity_vae}
\end{table*}

\begin{table*}[h]
\centering
{\small
\setlength{\tabcolsep}{1mm}
\begin{tabular}{l|l|l|l}
\hline
\textbf{Variants ($Q\%$)} & \textbf{DTLZ-1} & \textbf{DTLZ-2} & \textbf{DTLZ-3} \\
\hline
PMT-MOBO-DDPM-S (5\%) & 3.362e+04 (7.397e+02) & 3.091e+00 (1.630e-02) & 2.721e+04 (1.790e+03) \\
PMT-MOBO-DDPM (10\%) & \textbf{3.374e+04 (8.354e+02)} & \textbf{3.131e+00 (1.620e-02)} & \textbf{3.093e+04 (1.673e+03)} \\
PMT-MOBO-DDPM-L (20\%) & 3.367e+04 (6.563e+02) & 3.099e+00 (1.000e-02) & 2.888e+04 (2.341e+03) \\
\hline
\end{tabular}%
}
\caption{Sensitivity Analysis of Elite Ratio ($Q\%$) for DDPM. The proposed method demonstrates high stability, with the default configuration (10\%) consistently yielding peak Hypervolume.}
\label{tab:sensitivity_ddpm}
\end{table*}

\section{Experimental Setup}
\label{app:experiments}
The source code link: https://github.com/ambigeV/PEMOP.

\subsection{Generative Model Training}
\label{sec:genai}
We train two types of conditional generative models: a variational autoencoder (VAE) and a denoising diffusion probabilistic model (DDPM), as introduced in the main text. 
For the VAE, the encoder takes the concatenated input $[\mathbf{x},\mathbf{c}]$ and projects it through a single feed-forward layer to obtain a $d_{\rm lat}$-dimensional hidden representation. From this representation, two parallel linear layers produce the $d_{\rm man}$-dimensional mean and log-variance vectors. The decoder mirrors this structure: it first applies a feed-forward layer to the conditional latent code to yield another $d_{\rm lat}$-dimensional feature, then maps the result back to the decision space $\mathcal{X}$ via a linear layer. In our experiments, we set $d_{\rm lat}=D$ for the lack of training data, and $d_{\rm man}=2$ since each conditional Pareto Front lies on a one- or two-dimensional manifold. As for the loss function on the ELBO for VAE, instead of imposing the same weight for both reconstruction error and KL-divergence, we apply a weight $\beta=0.001$ to the KL-divergence so that we can achieve high-fidelity reconstruction quality.
Also, we simplify the conditional prior by replacing the $p(z|\boldsymbol{c})$ with the standard normal distribution $\mathcal{N}(0,\mathbf{I})$.

For the DDPM, we use a simple MLP architecture that takes the concatenated input $[\mathbf{x}, t, \boldsymbol{c}]$, where $t$ is the timestep, and $\mathbf{c}$ is the conditioning vector. 
The network consists of 4 fully connected layers with hidden dimension 128, following the structure of~\cite{Diffuse-BBO}. We employ a linear noise schedule with $\beta$ values ranging from 0.0001 to 0.02 over 1000 time steps. The model is trained to predict the noise $\boldsymbol{\epsilon}$ added during the forward diffusion process using standard MSE loss. During training, we randomly sample timesteps $t \sim \text{Uniform}(0, T-1)$ and compute the loss $\mathcal{L} = \|\boldsymbol{\epsilon} - \boldsymbol{\epsilon}_\phi(\mathbf{x}_t, t, \mathbf{c})\|_2^2$. For sampling, we use the standard DDPM reverse process, starting from Gaussian noise and iteratively denoising over all timesteps. We apply gradient clipping with a maximum norm of 1.0 for training stability and clamp the generated samples to the $[0,1]$ range to handle boundary constraints.
To accelerate this DDPM sampling process, we adopt the widely applied dpm-solver~\cite{lucheng} for generative solution sampling during and after the optimization process.

We employ Adam optimizer~\cite{kingma2017adammethodstochasticoptimization} for both models, and the learning rate is set to 0.1 for VAE and 0.001 for DDPM.

\subsection{GP Model Training}
\label{sec:gp}
For the Gaussian Process model, we implement a composite kernel GP that decomposes the input space into decision variables $\mathbf{x} \in \mathbb{R}^D$ and task parameters $\boldsymbol{\theta} \in \Theta \subset \mathbb{R}^{V}$. The model uses exact GP inference with a Gaussian likelihood, and the related hyperparameters are trained using the exact marginal log-likelihood optimized by Adam~\cite{kingma2017adammethodstochasticoptimization}.

The composite kernel is defined as the product of two kernel functions:
\[
\widetilde{\kappa}\bigl((\mathbf{x},\boldsymbol{\theta}),(\mathbf{x}',\boldsymbol{\theta}')\bigr) = \kappa_{\text{dec}}(\mathbf{x},\mathbf{x}') \cdot \kappa_{\text{task}}(\boldsymbol{\theta},\boldsymbol{\theta}')
\]
where $\kappa_{\text{dec}}$ is an isotropic RBF kernel for decision variables with lengthscale constrained to the interval $[0.1, 2.5]$, and $\kappa_{\text{task}}$ is an ARD RBF kernel for task parameters, allowing each task dimension to have its own lengthscale parameter. The overall covariance function is wrapped by a scale kernel to scale the product of the two kernels. This composite kernel is partially inspired by the kernel function in~\cite{offline-contextual}. The GP-based source codes are partially supported by GPytorch~\cite{Gpytorch}.
To optimize the acquisition function, we employed Adam optimizer~\cite{kingma2017adammethodstochasticoptimization} and the learning rate is set to 0.01.

\subsection{Data Collection for Generative Models}
\label{sec:data_collection}

To ensure the conditional generative model learns a high-fidelity representation of the Pareto manifold, we employ an adaptive data collection strategy. This method extracts high-quality solutions from the complete evaluation history while dynamically adjusting the resolution of preference vectors based on the complexity of the current approximate Pareto Front.

First, we accumulate the dataset of evaluated solutions from the optimization history to form a candidate pool, denoted as $\mathcal{D}_{pool}$. To determine the appropriate resolution for the scalarization weight vectors, we perform non-dominated sorting on $\mathcal{D}_{pool}$ to identify the non-dominated set (Rank 1, denoted as $\mathcal{F}_1$) and the second front (Rank 2, denoted as $\mathcal{F}_2$, the non-dominated set with Rank 1 solutions removed). We calculate the combined size of these high-quality fronts as $N_{front} = |\mathcal{F}_1| + |\mathcal{F}_2|$.

Next, we adaptively determine the number of uniformly sampled scalarization weight vectors, $W$, expected to adequately cover the objective space. The value of $W$ is derived from the front size $N_{front}$ and the objective dimension $M$, following the heuristic $W = (N_{front})^{M-1}$. Therefore, we increase the number of scalarization directions for a higher-dimensional objective space. We then uniformly sample $W$ weight vectors $\{\lambda_i\}_{i=1}^{W}$ in the objective space.

Finally, we construct the training dataset $\mathcal{D}^{(gen)}$ by scalarizing the entire candidate pool $\mathcal{D}_{pool}$. For each sampled weight vector $\lambda_i$, we compute the scalarized values of all solutions in $\mathcal{D}_{pool}$ and extract the top $Q\%$ elite solutions. This strategy is expected to make the generative model trained on a distribution that is performance-oriented via elite selection based on scalarized utility and can also diversely spread across the Pareto Front through the adaptive resolution of $W$ directions.

\subsection{Other Detailed Settings}
\label{sec:details}
In this work, to convert multi-objective vectors into single-objective scalar values, we instantiate $\mathfrak{s}_{\boldsymbol{\lambda}}(\cdot)$ for all methods using a recently proposed \textit{Hypervolume scalarization} function~\cite{hv-scalar}.
In the alternating optimization loop, the number of the sampled candidate solutions via the generative model per task $N_{gen}$ is set to 5,000, which is a common setting as parallel works blending inverse model and expensive multi-objective optimization~\cite{moo-finv}.

\section{Detailed Experimental Results}

\subsection{Detailed Average Hypervolume Comparison}
\label{sec:comparison}
Due to page limit in the main text, we showcase the detailed optimization results in Table~\ref{tab:results}. 
The results imply that the proposed generative optimizers, PMT-MOBO-VAE and PMT-MOBO-DDPM, generally achieve superior convergence and solution quality compared to single-task, multi-task, and generative baselines across distinct tasks.

We first compare our method against single-task approaches, including standard MOBO methods (ST-MOBO, qPAREGO, qEHVI) and Pareto Set Learning variants (PSL-MOBO). 
As indicated by the statistical tests, our methods significantly outperform standard baselines such as ST-MOBO and qPAREGO in all scenarios (denoted with $^{- -}$). 
While the acquisition-based method qEHVI remains competitive on specific landscapes, achieving the top performance on the MAGNETIC problem, PMT-MOBO-DDPM surpasses it on 6 out of 7 benchmarks. 
This suggests that while specialized acquisition functions are powerful in specific problem sets, generative modeling of the solution manifold offers more consistent performance across diverse applications.

We further evaluate the benefit of our generative solution sampling against existing multi-task MOBO methods, specifically OCMOBO and PPSL-MOBO, as well as our own ablation PMT-MOBO (without generative sampling). 
PMT-MOBO-DDPM consistently outperforms OCMOBO and the pure acquisition ablation (PMT-MOBO), confirming that simply utilizing data pooling is insufficient for approximating complex Pareto Fronts. 
The addition of the generative proposal distribution proves critical for search in complicated spaces. 
Against PPSL-MOBO, a strong multi-task baseline that can perform well on DTLZ-3, our DDPM variant achieves higher Hypervolume on the majority of real-world tasks (LAMP, SOLAR, UAV), showcasing the superior flexibility of diffusion models in capturing complex Pareto Fronts compared to others.

A critical comparison is against CDM-PSL, which also utilizes a diffusion model but operates in a single-task setting. The results indicates a substantial performance gap, with PMT-MOBO-DDPM significantly outperforming CDM-PSL in every benchmark (e.g., achieving $3.37\text{e}4$ versus $1.59\text{e}4$ on DTLZ-1). 
This validates our core hypothesis regarding inter-task synergy: by fully utilizing data and search information across related tasks to train a conditional generative model, we overcome the data scarcity that can hamper single-task generative models like CDM-PSL, allowing our optimizer to learn robust inverse maps even under tight evaluation budgets. 
In summary, PMT-MOBO-DDPM possesses the most robust generalization capabilities, achieving the best average Hypervolume on 6 out of 7 benchmarks.

\subsection{Structural Differences in Amortizing Multi-Objective Optimization: Ours vs. PPSL-MOBO}
\label{sec:novelty}
We recognize PPSL-MOBO as a well-designed parametric multi-objective optimizer based on PSL-MOBO~\cite{moo-psl}. 
However, a fundamental distinction lies in the architectural assumptions of the inverse model used for amortization. 
PPSL-MOBO utilizes a deterministic hypernetwork to learn a functional mapping from task-preferences to solutions~\cite{ppsl}. 
This approach implicitly assumes a \textit{bijective} (one-to-one) relationship, where each preference vector is mapped to a single, unique solution in the decision space.

In practice, this deterministic structure forces the model to approximate the Pareto set as a scalar function of the inputs. 
When the underlying Pareto manifold is wiggly or highly irregular, the hypernetwork might struggle to represent potential multiple solution modes for a single preference, often resulting in an over-smoothed Pareto Front. 
As shown in our results for the SOLAR and LAMP benchmarks in Figure~\ref{fig: PF-Comparison-PPSL}, this leads to a limited Pareto Front spread, where the optimizer fails to capture the diversity and extreme trade-offs of the true Pareto Front.

Differently, our PMT-MOBO-X framework treats the mapping as a distributional learning problem, enabling a potential one-to-many relationship. 
By modeling the density of elite solutions via VAEs or diffusion models, the framework can represent multiple solution modes for a single task-preference query. 
As verified by the broader coverage in Figure~\ref{fig: PF-Comparison-DDPM} and the superior Hypervolume results in Table~\ref{tab:results}, this allows our method to discover more expansive and complex Pareto Front shapes, providing the flexibility necessary for robust amortized multi-objective optimization.

\subsection{Ablation Studies}
\label{sec:ablation}

Tables~\ref{tab:ablation_vae} and \ref{tab:ablation_ddpm} present the ablation results for VAE and DDPM integrations, respectively. 
We study the impact of the alternating loop by comparing the full method against two decoupled baselines: pure acquisition search (PMT-MOBO w/o \texttt{GENERATIVE}) and pure generative sampling (PLAIN-VAE/DDPM).
The results for the VAE variant (Table~\ref{tab:ablation_vae}) clearly illustrate the limitations of isolated components.
PLAIN-VAE exhibits a high standard deviation across benchmarks (e.g., $7.56\text{e}3$ on DTLZ-1), indicating that while pure generative sampling can occasionally "get lucky" and hit high-quality solutions, it suffers from inherent stochasticity during the optimization process and lacks the precision to consistently refine the Pareto Front.
Conversely, pure acquisition search often converges prematurely to local optima, failing to explore promising but unseen regions of the design space.
By coupling these components, PMT-MOBO-VAE leverages generative sampling to enhance candidate diversity while utilizing the alternating acquisition search to lower the variance of pure sampling. 
This synergy results in statistically superior convergence and stability across all tested problems.

\subsection{Sensitivity Analysis}
\label{sec:sensitivity}

We investigate the sensitivity of the elite ratio parameter $Q$.
The trend is most pronounced in the VAE sensitivity analysis shown in Table~\ref{tab:sensitivity_vae}.
\begin{itemize}
\item \textbf{Large $Q$ (20\%):} Increasing the elite ratio dilutes the quality of the training set. This reduces the selective pressure, causing the generative model to learn the distribution of sub-optimal, low-quality solutions rather than focusing on the high-performing elite regions. Consequently, performance drops on DTLZ-1 and DTLZ-3.
\item \textbf{Small $Q$ (5\%):} Conversely, an overly strict selection criterion limits the training data availability. While this theoretically sharpens exploitation, in practice, it prevents the model from learning a well-behaved latent structure, incurring large variance and instability (e.g., higher standard deviation on DTLZ-1).
\item \textbf{Optimal $Q$ (10\%):} The default setting strikes an effective balance, providing sufficient data for stable training while maintaining high selective pressure to guide the model toward the Pareto Front.
\end{itemize}
While the DDPM variant (Table~\ref{tab:sensitivity_ddpm}) shows stable results with respect to the parameter changes, the underlying trade-off remains consistent: $Q=10\%$ consistently yields the most robust Hypervolume performance.

\begin{table*}[h]
\centering
{\small
\setlength{\tabcolsep}{1mm}
\begin{tabular}{l|l|l}
\hline
\textbf{Component} & \textbf{PMT-MOBO-VAE} (s) & \textbf{PMT-MOBO-DDPM} (s) \\
\hline
GP Training & $7.92 \pm 2.64$ & $7.19 \pm 2.25$ \\
Acquisition Optimization & $6.10 \pm 0.34$ & $5.90 \pm 0.31$ \\
Generative Model Training & $0.55 \pm 0.26$ & $0.39 \pm 0.15$ \\
Generative Solution Sampling & $6.08 \pm 0.28$ & $6.03 \pm 0.50$ \\
\hline
\end{tabular}}
\caption{Runtime Overhead Analysis on DTLZ-2 ($T=50$). The table reports the wall-clock time (Mean $\pm$ Std) for each algorithmic component per active iteration. Acquisition Optimization occurs only in \texttt{ACQUISITION} mode, while generative model training and solution sampling occur only in \texttt{GENERATIVE} mode. GP Training occurs in every iteration.}
\label{tab:runtime_analysis}
\end{table*}

\subsection{Computational Overhead}
A potential concern with using generative models into the optimization loop is the computational cost of iterative retraining.
To address this, we profiled the wall-clock time of each algorithmic component on the DTLZ-2 benchmark over $T=50$ iterations.
As detailed in Table~\ref{tab:runtime_analysis}, the cost of generative model training is negligible ($< 0.6$s per iteration), orders of magnitude faster than the standard Gaussian Process (GP) training ($\approx 7.9$s).
Furthermore, the generative solution sampling ($\approx 6.0$s) incurs a computational cost comparable to standard acquisition optimization ($\approx 6.1$s).
Since our alternating framework swaps between acquisition and generative modes rather than running them simultaneously, the total runtime per iteration remains consistent with standard MOBO methods. 
This confirms that our framework enhances optimization performance without imposing a prohibitive computational burden.

\subsection{Pareto Fronts Generated via Inverse Models}
\begin{figure*}[th]
\centering
\includegraphics[width=0.88\textwidth]{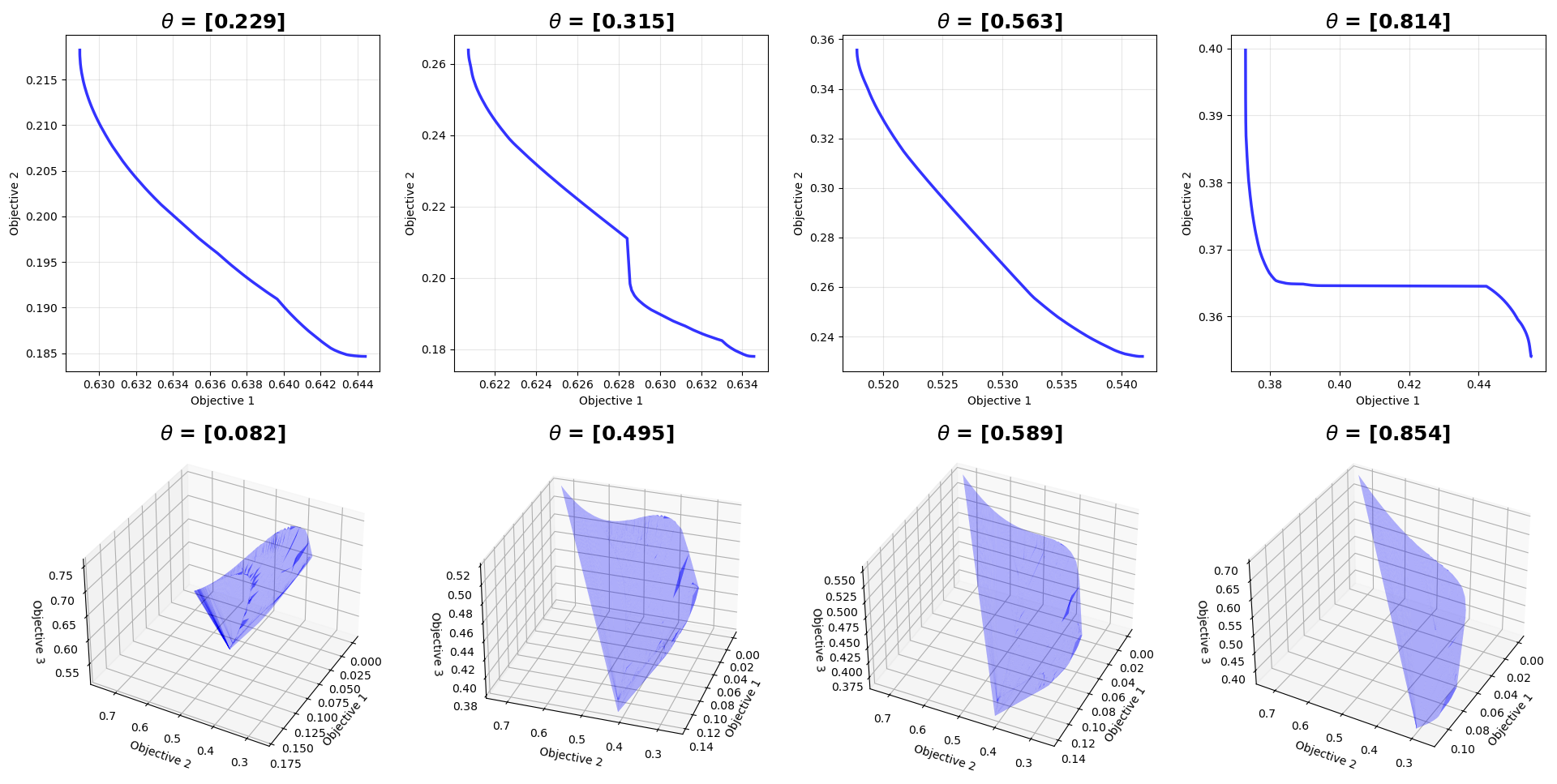}
\caption{Approximated Pareto Front for unseen EMOP tasks in solar rooftop generation (top) and lamp generation (bottom) by PSL-MOBO.}
\label{fig: PF-Comparison-PSL}
\end{figure*}

\begin{figure*}[th]
\centering
\includegraphics[width=0.88\textwidth]{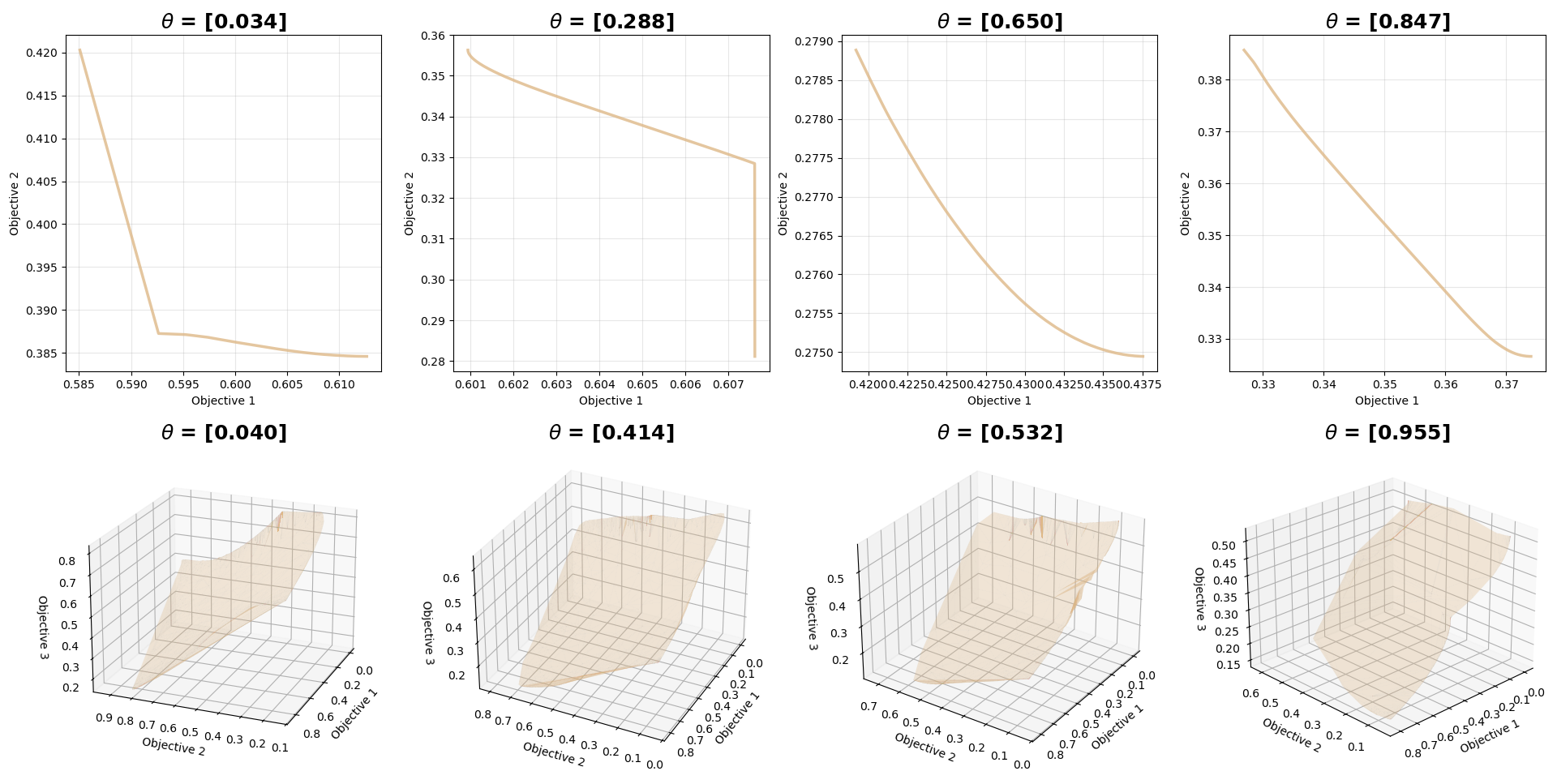}
\caption{Approximated Pareto Front for unseen EMOP tasks in solar rooftop generation (top) and lamp generation (bottom) by PPSL-MOBO.}
\label{fig: PF-Comparison-PPSL}
\end{figure*}

\begin{figure*}[th]
\centering
\includegraphics[width=0.88\textwidth]{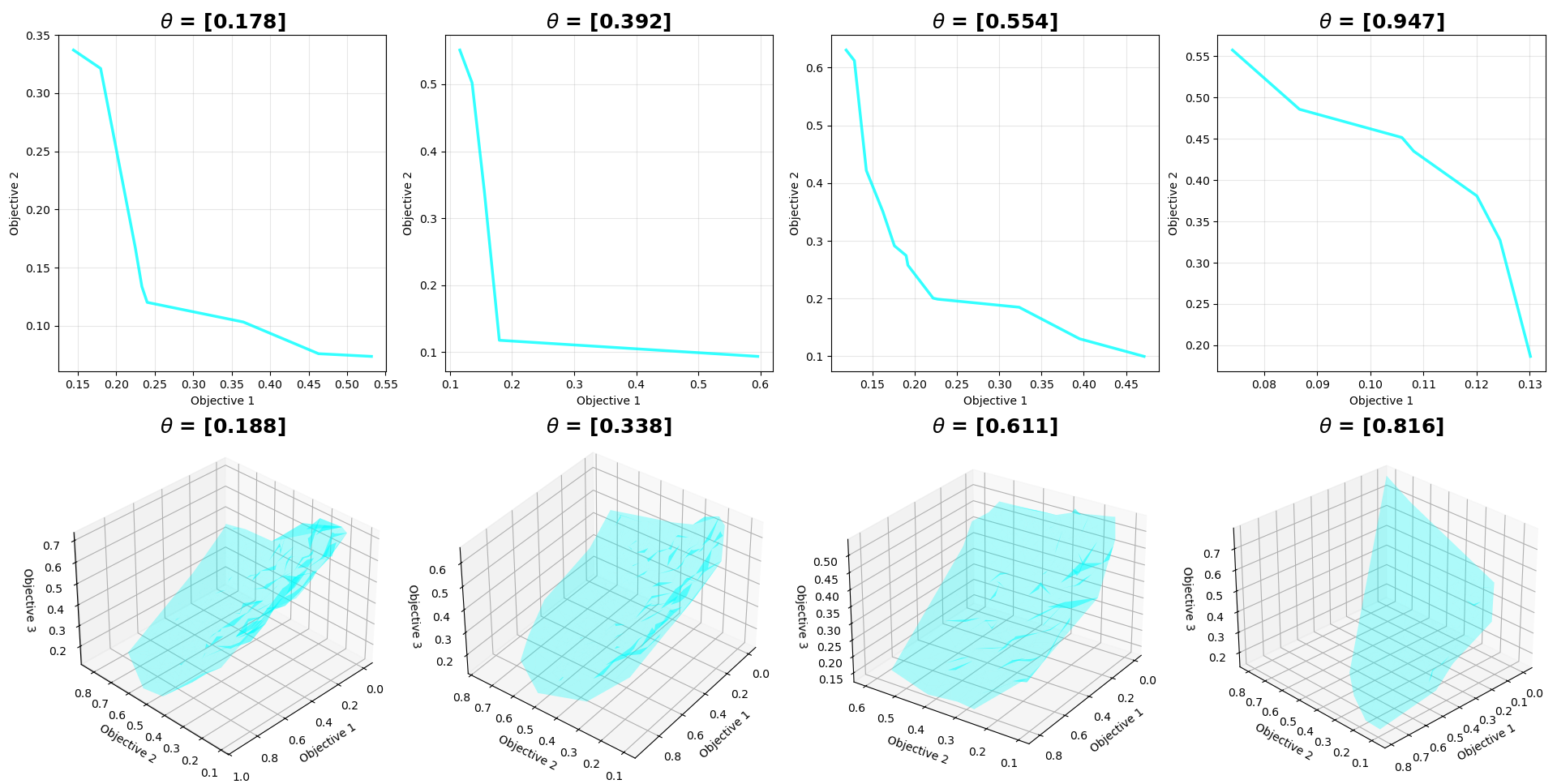}
\caption{Approximated Pareto Front for unseen EMOP tasks in solar rooftop generation (top) and lamp generation (bottom) by CDM-PSL.}
\label{fig: PF-Comparison-CDM}
\end{figure*}

\begin{figure*}[th]
\centering
\includegraphics[width=0.88\textwidth]{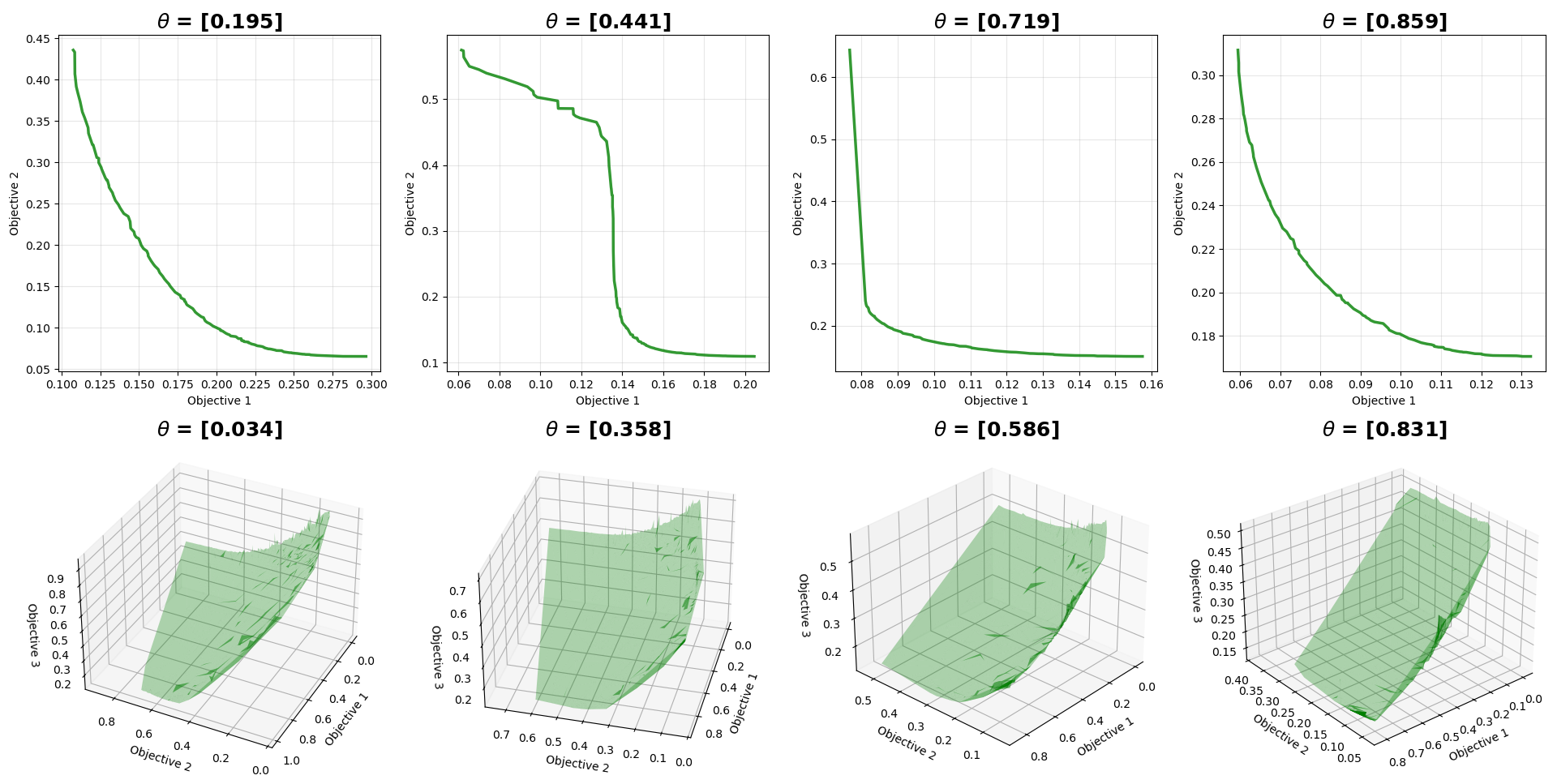}
\caption{Approximated Pareto Front for unseen EMOP tasks in solar rooftop generation (top) and lamp generation (bottom) by PMTO-MOBO-VAE.}
\label{fig: PF-Comparison-VAE}
\end{figure*}

\begin{figure*}[th]
\centering
\includegraphics[width=0.88\textwidth]{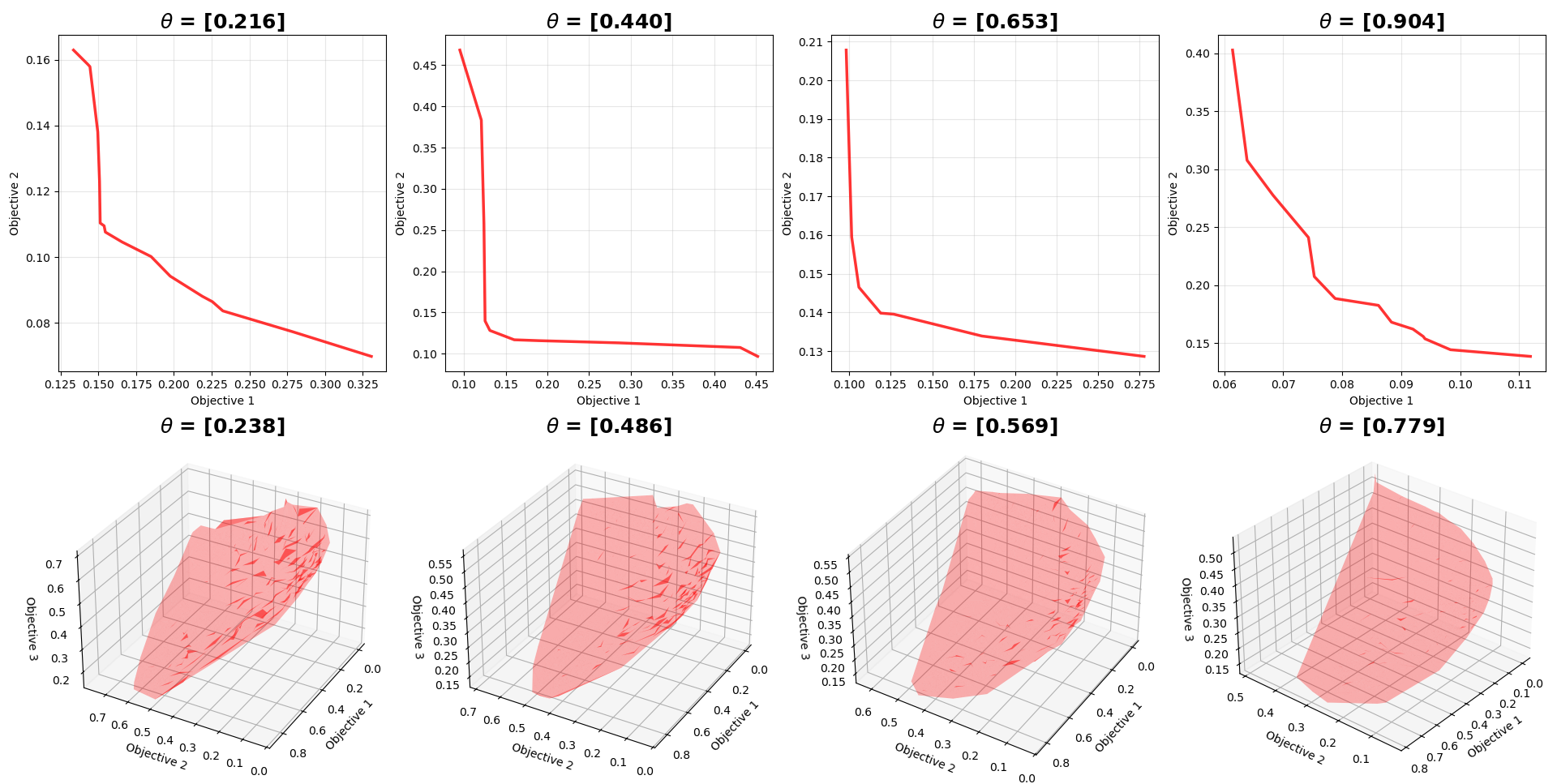}
\caption{Approximated Pareto Front for unseen EMOP tasks in solar rooftop generation (top) and lamp generation (bottom) by PMT-MOBO-DDPM.}
\label{fig: PF-Comparison-DDPM}
\end{figure*}

In this subsection, we showcase those exemplar Pareto Fronts generated by the inverse models obtained by the algorithm PSL-MOBO in Figure~\ref{fig: PF-Comparison-PSL}, CDM-PSL in Figure~\ref{fig: PF-Comparison-CDM}, PPSL-MOBO in Figure~\ref{fig: PF-Comparison-PPSL}, PMT-MOBO-VAE in Figure~\ref{fig: PF-Comparison-VAE}, and PMT-MOBO-DDPM in Figure~\ref{fig: PF-Comparison-DDPM}.

\section{Related Works}
\label{app:code}
\subsection{Inverse Modeling for Multi-Objective Optimization} 
Learning direct inverse maps from task–preference inputs to Pareto-optimal solutions enables flexible preference articulation for a family of expensive multi-objective optimization problems without expensive re-optimization. 
Early work treated the Pareto Front as a smooth manifold to be approximated \cite{Flemming}. Subsequently, \cite{7018980} first integrated inverse modeling into the optimization loop itself, followed by Pareto Set Learning for EMOP \cite{moo-psl}, which employed inverse predictions for both post-hoc preference articulation and active search assistance. 
Recognizing the inherent knowledge in discovered Pareto-optimal solutions, subsequent work leveraged inverse models to streamline related multi-objective optimization~(MOO) tasks: providing initial solutions for dynamic MOO~\cite{9440867}, transferring quality solutions to related MOO tasks~\cite{inverse-2}, or directly building inverse multi-task models \cite{moo-finv} to learn multiple Pareto Sets or Pareto Fronts simultaneously.

Our work operates in a fundamentally different paradigm by mapping the joint space of task and preference parameters directly to solution space. Rather than handling discrete sets of related MOO tasks, we parameterize a family of infinitely many MOO tasks and enable zero-shot generalization through task-aware exploration coupled with conditional generative inverse sampling.

\subsection{Few-Shot Bayesian Optimization} 
Few-shot BO aims to rapidly adapt optimization configurations to new instances with minimal evaluations. Existing approaches focus on transferring optimizer components: multi-task GP kernels model inter-task relationships \cite{swersky2013multi}, meta-learned deep kernels configure GP surrogates from metadata of other optimization tasks~\cite{wistuba2021fewshotbayesianoptimizationdeep}, reinforcement learning adapts acquisition functions \cite{NEURIPS2021_3fab5890}, and meta-priors enable transfer across heterogeneous spaces \cite{JMLR:v25:23-0269,fan2024transfer}. 

Our approach differs by targeting solution space regularity rather than adjusting optimizer configuration. We learn a conditional generator over Pareto-optimal candidates across the task space and couple it with task-aware GPs. The conditional generative model provides structured solution priors for novel tasks while GPs maintain principled uncertainty quantification. This design transfers solution manifold knowledge directly rather than indirectly sharing the surrogate (hyper-)parameters.

\begin{algorithm}[h]
\caption{Workflow of ST-MOBO and PMT-MOBO}
\label{alg:pmobo-vs-ucb}
\textbf{Input}: Tasks $\{\boldsymbol{\theta}_k\}_{k=1}^K$, objectives $\{f_m\}_{m=1}^M$, preference distribution $\Lambda$, method $\mathcal{A} \in \{\texttt{ST-MOBO}, \texttt{PMT-MOBO}\}$.\\
\textbf{Output}: Final datasets $\{\mathcal{D}_k\}_{k=1}^K$
\begin{algorithmic}[1]
\FOR{$t = 1$ \TO $T$}
    \STATE Sample preference vector $\boldsymbol{\lambda} \sim \Lambda$
    \FOR{each task $\boldsymbol{\theta}_k$}
        \IF{$\mathcal{A} = \texttt{ST-MOBO}$}
            \STATE Select $\mathbf{x}_k^{(t)}$ using~(\ref{eq:acquisition}) and (\ref{eq:ucb-mobo})
        \ELSE
            \STATE Select $\mathbf{x}_k^{(t)}$ using~(\ref{eq: task-acquisition}) and (\ref{eq:pmt-mobo-ucb})
        \ENDIF
        \STATE $\mathcal{D}_k \leftarrow \mathcal{D}_k \cup \{\bigl(\mathbf{x}_k^{(t)}, \boldsymbol{\theta}_k, F(\mathbf{x}_k^{(t)}, \boldsymbol{\theta}_k)\bigr)\}$
    \ENDFOR
    \FOR{each objective $f_m$}
        \IF{$\mathcal{A} = \texttt{ST-MOBO}$}
            \FOR{each task $\boldsymbol{\theta}_k$}
                \STATE $\mathcal{D}_k^{(m)} \leftarrow \{\bigl(\mathbf{x}_{k,i}, f_m(\mathbf{x}_{k,i}, \boldsymbol{\theta}_k)\bigr)\}_{i=1}^{N_k}$ 
                \STATE Train $\mathcal{GP}_m$ on $\mathcal{D}_k^{(m)}$
            \ENDFOR
        \ELSE
            \STATE $\mathcal{D}_{\text{joint}}^{(m)} \leftarrow \bigcup_k \{\bigl(\mathbf{x}_{k,i}, \boldsymbol{\theta}_k, f_m(\mathbf{x}_{k,i}, \boldsymbol{\theta}_k)\bigr)\}_{i=1}^{N_k}$ 
            \STATE Train $\mathcal{GP}_{TA,m}$ on $\mathcal{D}_{\text{joint}}^{(m)}$
        \ENDIF
    \ENDFOR
\ENDFOR
\STATE \textbf{return} $\{\mathcal{D}_k\}_{k=1}^K$
\end{algorithmic}
\end{algorithm}

\section{Workflow of ST-MOBO and PMT-MOBO}
\label{sec:algo}
Due to the page limit, we detaile the comparative workflow of ST-MOBO and PMT-MOBO here in \textbf{Algorithm \ref{alg:pmobo-vs-ucb}}.

\end{document}